\documentclass[times,twocolumn,final]{elsarticle}

\usepackage{cag}
\usepackage{framed,multirow}

\usepackage{amssymb}
\usepackage{latexsym}
\usepackage{subcaption}
\usepackage{arydshln}
\usepackage{afterpage}
\usepackage{graphicx}

\usepackage{url}
\usepackage{xcolor}
\definecolor{newcolor}{rgb}{.8,.349,.1}

\usepackage{hyperref}

\usepackage[switch,pagewise]{lineno} 

\usepackage{amsmath}

\journal{Computers \& Graphics}

\begin{document}

\verso{Preprint Submitted for review}

\begin{frontmatter}

\title{No-reference Geometry Quality Assessment for Colorless Point Clouds via List-wise Rank Learning}%








\author[mymainaddress]{Zheng Li}
\ead{1315571938@qq.com}

\author[mymainaddress]{Bingxu Xie}
\ead{2580801219@qq.com}

\author[mymainaddress]{Chao Chu}
\ead{979953220@qq.com}

\author[mysecondaryaddress]{Weiqing Li}
\ead{li\_weiqing@njust.edu.cn}

\author[mymainaddress]{Zhiyong Su\corref{mycorrespondingauthor}}
\cortext[mycorrespondingauthor]{Corresponding author}
\ead{su@njust.edu.cn}

\address[mymainaddress]{School of Automation, Nanjing University of Science and Technology, Nanjing, Jiangsu Province 210094, P.R. China}

\address[mysecondaryaddress]{School of Computer Science and Engineering, Nanjing University of Science and Technology, Nanjing, Jiangsu Province 210094, P.R. China}

\received{\today}

\begin{abstract}
Geometry quality assessment (GQA) of colorless point clouds is crucial for evaluating the performance of emerging point cloud-based solutions (e.g., watermarking, compression, and 3-Dimensional (3D) reconstruction). 
Unfortunately, existing objective GQA approaches are traditional full-reference metrics, whereas state-of-the-art learning-based point cloud quality assessment (PCQA) methods target both color and geometry distortions, neither of which are qualified for the no-reference GQA task.
In addition, the lack of large-scale GQA datasets with subjective scores, which are always imprecise, biased, and inconsistent, also hinders the development of learning-based GQA metrics. 
Driven by these limitations, this paper proposes a no-reference geometry-only quality assessment approach based on list-wise rank learning, termed LRL-GQA, which comprises of a geometry quality assessment network (GQANet) and a list-wise rank learning network (LRLNet).
The proposed LRL-GQA formulates the no-reference GQA as a list-wise rank problem, with the objective of directly optimizing the entire quality ordering.
Specifically, a large dataset containing a variety of geometry-only distortions is constructed first, named LRL dataset, in which each sample is label-free but coupled with quality ranking information.
Then, the GQANet is designed to capture intrinsic multi-scale patch-wise geometric features in order to predict a quality index for each point cloud.
After that, the LRLNet leverages the LRL dataset and a likelihood loss to train the GQANet and ranks the input list of degraded point clouds according to their distortion levels.
In addition, the pre-trained GQANet can be fine-tuned further to obtain absolute quality scores.
Experimental results demonstrate the superior performance of the proposed no-reference LRL-GQA method compared with existing full-reference GQA metrics.
The source code can be found at : \href{https://github.com/VCG-NJUST/LRL-GQA}{https://github.com/VCG-NJUST/LRL-GQA}.
\end{abstract}

\begin{keyword}
\KWD Geometric Point Cloud Quality Assessment \sep List-wise Ranking Learning \sep No-Reference Geometry-Only Quality Metrics
\end{keyword}

\end{frontmatter}


\section{Introduction} 
\label{sec:introduction}
Geometry quality assessment (GQA) of colorless point clouds refers to the process of taking a colorless point cloud as input and producing some form of geometry quality estimate as output, which plays an essential role in numerous point cloud-based applications, such as denoising, watermarking, acquisition, and compression.
Nowadays point clouds have emerged as an important and practical representation of 3-dimensional (3D) objects and surrounding environments, thanks to the rapid development of geometric sensing techniques.
Unfortunately, point clouds are subject to a variety of geometric distortions that influence the geometry quality of the data throughout the acquisition to rendering pipeline.
Therefore, in order to automatically estimate the extent of geometric deterioration produced by these processes, objective geometry quality measurements are required. 
These metrics can subsequently be employed in the design and optimization of point cloud-based algorithms and systems.

To date, numerous objective point cloud quality assessment (PCQA) methods have been developed for both colorless and colored point clouds \citep{Alexioue18,Dumice18,Fretesh19,Nehmey19,niuyz19,Lin22, diniz2022point}, but there are still some limitations in the objective GQA.
\textbf{First of all}, existing objective GQA approaches are traditional full-reference metrics, such as point-to-point (Po2Po) \citep{Rufaelm16}, point-to-distribution (Po2D) \citep{Javaheria20a}, point-to-plane (Po2Pl) \citep{Tiand17}, and plane-to-plane (Pl2Pl) \citep{Alexioue18icme}.
In practice, however, the reference point clouds are typically unavailable, e.g., denoising and reconstruction. 
Additionally, existing learning-based PCQA methods all target both color and geometry distortions \citep{wuxj21,Yangq21}.
Since the color attribute plays a crucial part in human perception throughout the process of gathering subjective ratings, such as mean opinion score (MOS) which represents human subjective perception quality through large-scale human scoring, these algorithms cannot be employed directly for the GQA task.
\textbf{Second}, existing learning-based PCQA approaches primarily focus on assigning absolute quality scores to distorted point clouds.
They need sufficient data with accurate ground-truth quality scores for training, which usually come from subjective experiments that are time-consuming, expensive, and require strict control conditions.
Furthermore, objective scores are usually imprecise, biased, and inconsistent \citep{Gaof15, mkd17, Nehmey19, Hub19, oufz21}.
For example, the scores provided by different candidates or even the same candidate at different periods for the same point clouds are typically different.
\textbf{Third}, existing publicly available PCQA datasets with MOS include both geometry and color distortions, such as coloured point cloud database (CPCD) 2.0 \citep{Hual21}, SJTU point cloud quality assessment database (SJTU-PCQA) \citep{Yangq21}, SIAT point cloud quality database (SIAT-PCQD) \citep{wuxj21}, Waterloo point cloud database (WPC) \citep{liuq22}, large scale point cloud quality assessment database (LS-PCQA) \citep{Liuyp22} and broad quality assessment of static point clouds in compression scenario (BASICS) \citep{BASICS2024}.
And, the existing largest colorless geometry point cloud dataset (G-PCD) \citep{Alexioue17} only has 5 reference point clouds and 40 distorted instances, which are not sufficient to derive learning-based GQA metrics with high generalization ability.
Therefore, developing no-reference learning-based objective GQA methods for colorless point clouds is still an open and challenging problem.

Inspired by existing rank learning-based image quality assessment (IQA) methods \citep{Gaof15,Mal16,Liuxl17,Mengxd21,Fuzq22,Wangxj22}, this paper explores to formulate the GQA as a quality-based ranking problem and proposes a no-reference point-based GQA method based on deep list-wise rank learning (LRL), called LRL-GQA. 
To alleviate the shortage of large MOS-aware datasets with geometry-only distortions, this paper constructs a new ranking dataset, named LRL dataset.
The LRL dataset provides the quality ranking order of a list of degraded point clouds as ground-truth, which is much easier than manually assigning absolute scores to each distorted point cloud.
The proposed LRL-GQA method consists of a geometry quality assessment (GQANet) network and a list-wise rank learning network (LRLNet). 
The GQANet is designed to encode patch-wise geometric features in multiple scales and then predict a quality index for each input point cloud.
The LRLNet adopts the list-wise minimizing likelihood estimation (listMLE) \cite{xia2008listwise} as the loss function to train the GQANet to rank the input list of distorted point clouds based on the LRL dataset.
In addition, the GQANet is demonstrated to be capable of calculating absolute quality scores after fine-tuning on a new pseudo mean opinion score dataset, named LRL-PMOS, in which each sample is assigned with a pseudo quality score as its substitute of MOS.
The main contributions of this paper are as follows:
\begin{enumerate}
	\item[1)] This paper formalizes the no-reference GQA as a list-wise rank learning problem, which directly learns to rank a list of distorted point clouds generated from the same source by the same geometric distortion at different distortion levels.
	To the best of our knowledge, this is the first deep list-wise rank learning-based geometry-only quality assessment method. 
	
	\item[2)] A geometry quality assessment network (GQANet) is designed to encode patch-wise quality-aware features in multi-scale to predict the quality index of each point cloud.
	The pre-trained GQANet is capable of predicting absolute quality scores after fine-tuning.
	
	\item[3)] A large-scale ranking dataset called LRL is generated, which contains 52200 geometrically distorted point clouds derived from 200 original colorless reference point clouds, each with 26 types of distortions at 10 distortion levels.
	Based on the LRL dataset, a large-scale dataset with pseudo mean opinion score named LRL-PMOS is built, in which each sample is assigned with a pseudo quality score calculated by an existing outstanding full-reference GQA metric.
\end{enumerate}

The rest of this paper is organized as follows. 
Section \ref{sec:Related Works} reviews the studies about point cloud quality assessment and rank learning-based IQA.
Section \ref{sec:dataset} introduces the construction of LRL dataset. 
Section \ref{sec:Proposed Method} describes the proposed geometry quality assessment framework based on list-wise learning in detail. 
Section \ref{sec:Experimental} presents experimental studies to demonstrate the state-of-the-art performance of the proposed framework. 
Section \ref{sec:Conclusion} concludes the paper.

\section{Related Works} 
\label{sec:Related Works}

This section first reviews objective PCQA approaches and subjective PCQA datasets.
Then, typical rank learning-based IQA methods are discussed.

\subsection{Objective PCQA Methods}
\label{subsec:spcqa}

Objective PCQA methods \citep{liuq22,oufz21} can be classified into geometry-only and general-purpose approaches from the perspective of application scope.
Geometry-only PCQA methods evaluate the geometry quality of point clouds based on geometry attributes alone, whereas general-purpose PCQA approaches take geometry and color attributes into consideration simultaneously.

\subsubsection{Geometry-only PCQA Methods}
\label{subsec:fr_method_details}
Up to now, geometry-only PCQA methods  \citep{Alexioue18,Dumice18,Meynetg20,Yangq22,Dinizr21,liuq22} are all full-reference point-based approaches, which evaluate the geometry quality of point clouds based on geometry attributes in the 3D space directly.
These methods can be classified into Po2Po, Po2Pl, Po2D, and Pl2Pl metrics.

The Po2Po metric measures the degree of distortion by quantifying the Euclidean distances between corresponding points, such as Hausdorff distance (HD) \citep{Lavoueg10}, root mean square error (RMSE) \citep{Javaheria17a}, mean city-block distance (MCD) \citep{Zengj20}, peak signal to noise ratio (PSNR) \citep{Tiand17}, and Chamfer distance (CD) \citep{Zhangdb21}.
The Po2Pl metric \citep{Tiand17} improves over Po2Po by projecting the obtained Po2Po distances along the surface normal direction.
The Po2D \citep{Javaheria20a} distance exploits a new type of correspondence between two point clouds and employs the Mahalanobis distance to measure the distance between a point and a distribution of points in a limited point cloud region.
The Pl2Pl metric \citep{Alexioue18icme} is built on the angular similarity of tangent planes that correspond to associated points between the reference and the degraded point cloud.

\subsubsection{General-purpose PCQA Methods}

General-purpose PCQA methods cover point-based and projection-based approaches, which take both geometry and color attributes into consideration.

\textbf{Point-based approaches} predict absolute quality scores by evaluating the quality of point clouds directly in the point space. 
Viola et al. \citep{Violai20} extracted color statistics from both reference and degraded point clouds, and employed color histograms to drive objective metrics.
Meynet et al. \citep{Meynetg20} also selected several geometry-based and color-based features and combined them linearly by logistic regression.
In \citep{Yangq22}, Yang et al. proposed a metric, called GraphSIM, to predict the human perception of colored point clouds with superimposed geometry and color impairments.
They utilized graph signal processing to extract point cloud color gradients to yield robust quality prediction. 
Chai et al. \citep{Chai2024} introduced a potent no-reference point cloud quality evaluation technique, which combines visual and geometric characteristics, leading to a substantial improvement in the precision and applicability of the assessment.
They compared and combined the statistics of color and geometry information of the reference and original point cloud to estimate the perceived quality of the degraded point cloud.
The proposed coloured point cloud based on geometric segmentation and colour transformation (CPC-GSCT) \citep{Hual21} metric employed geometric segmentation and color transformation respectively to construct geometric and color features for estimating the point cloud quality.

In addition to the aforementioned full-reference point-based methods, no-reference point-based approaches only utilize degraded point clouds and do not require information about reference point clouds.
However, there is currently less research on no-reference quality assessment in the 3D space directly.
The proposed blind quality evaluator for colored point cloud based on visual perception (BQE-CVP) method \citep{Hual21cvp} utilized handcrafted features to develop a no-reference quality evaluator.  
It characterized the distortion of distorted point clouds from geometric, color, and joint perspectives.
Liu et al. \citep{Liuyp22} proposed a no-reference metric ResSCNN based on sparse convolutional neural network (SCNN) with residual connections, which is a type of deep learning architecture designed to process grid-like data, to accurately estimate the subjective quality of point clouds.
The proposed ResSCNN adopts a hierarchical feature extraction module to extract both geometry and color attributes of point clouds, which takes the entire point cloud as input.
Shan et al. \citep{Shanzy23} proposed a novel no-reference PCQA metric, called GPA-Net, to attentively extract perturbation of structure and texture via a multi-task graph convolutional network.
They introduced a coordinate normalization module to achieve the shift, scale, and rotation invariance.

\textbf{Projection-based methods}, which have dominated the field of PCQA, strive to project point clouds onto a set of 2-dimensional (2D) planes and then predict absolute quality scores using existing IQA techniques \citep{Zhangzc22, Alexioue19mex, wuxj21, Yangq21, Yang_2022_CVPR, liuq22, wang2024zoom}.

Most existing projection-based methods \citep{Zhangzc22} employ handcrafted features to characterize the distortion of degraded point clouds.
Alexiou et al. \citep{Alexioue19mex} investigated the impact of the number of viewpoints employed to assess the visual quality of point clouds and proposed to assign weights to the projected views based on interactivity information obtained during subjective evaluation experiments.
In \citep{wuxj21}, using a head-mounted display (HMD) with six degrees of freedom, Wu et al. proposed two projection-based objective quality evaluation methods: a weighted view projection based model and a patch projection-based model. 
He et al. \citep{Hezy21} projected the texture information and curvature of colored point clouds onto 2D planes, and extracted texture and geometric statistical features, respectively.
Their method combined both colored texture and curvature projection.
Yang et al.\citep{Yangq21} projected the 3D point cloud onto six perpendicular image planes of a cube to obtain both 2D texture and depth images to represent the photometric and geometric information of the original point cloud.
Then, they utilized features extracted from these images for objective metric development.
Yang et al. \citep{Yang_2022_CVPR} also leveraged the rich prior knowledge from the field of IQA and applied the domain adaptation to point cloud quality assessment.
Liu et al. \citep{liuq22} employed an attention mechanism and a variant of information content-weighted structural similarity to develop a novel objective model, which significantly outperforms existing metrics. 

\begin{table*}[!t]
	\centering
	\caption{Selected publicly accessible PCQA datasets.}
	\label{Tab:existing-dataset}
\resizebox{\textwidth}{!}{
	\begin{tabular}{lllll}
		\hline
		\multicolumn{1}{l}{Types}       & \multicolumn{1}{l}{Datasets} & \multicolumn{1}{l}{Reference samples} & \multicolumn{1}{l}{Distortion type}            & \multicolumn{1}{l}{Distorted samples} \\ \hline
		\multirow{2}{*}{Colorless} & G-PCD \citep{Alexioue17}     & 5   & Octree-puring, Gaussian noise & 40   \\
		& RG-PCD \citep{Alexioue18mex}   & 6   & Octree-puring  & 24  \\ \hline
		\multirow{6}{*}{Colored}   & CPCD 2.0 \citep{Hual21}   & 10    & G-PCC, V-PCC, Gaussian noise& 360    \\
		& SIAT-PCQD \citep{wuxj21}  & 20     & V-PCC   & 340       \\
		& SJTU-PCQA \citep{Yangq21}   & 10     & Octree, downsampling, color and geometry noise & 420    \\
		& WPC \citep{liuq22}   & 20    & Gaussian noise, dowsampling, G-PCC, V-PCC& 740                                   \\
		& LS-PCQA \citep{Liuyp22}     & 104      & G-PCC, V-PCC, color and geometry noise         & 22,568      \\ 
      & BASICS \citep{BASICS2024}     & 75      & V-PCC, G-PCC-Predlift, G-PCC-Raht, GeoCNN         & 1494      \\ 
  \hline
	\end{tabular}}
\end{table*}

Recently, deep learning-based feature extraction methods have also been introduced into projection-based approaches.
Liu et al. \citep{liuq21tcsvt} proposed a deep learning-based no reference point cloud quality assessment method, called PQA-Net, consisting of three modules: a multi-view-based joint feature extraction and fusion module, a distortion type identification module, and a quality vector prediction module.
The entire network is jointly trained for quality prediction. 
Tao et al. \citep{Taowx21} projected 3D point clouds into 2D color projection maps and geometric projection maps and designed a multi-scale feature fusion network to blindly evaluate the visual quality. 
However, the selection of projection directions may significantly influence the overall assessment performance \citep{Yangq22,Liuyp22}.
Consequently, the 3D-to-2D projection cannot characterize the 3D points distribution very well, resulting in information loss.

All in all, the majority of existing PCQA methods are specially designed for colored point clouds, which cannot be directly used for geometry-only quality assessment.
Moreover, current geometry-only PCQA methods are all full-reference and hand-crafted approaches.
In addition, all existing  PCQA methods are dedicated to predicting absolute quality scores of input point clouds.
This paper is dedicated to developing a deep list-wise rank learning framework for no-reference GQA in the 3D space.

\subsection{Subjective PCQA Datasets}

Subjective PCQA datasets are widely used to train, validate, and test learning-based PCQA methods.
Existing datasets can be classified into colored and colorless PCQA datasets, according to the color attribute of their original point clouds, while also containing various types of distortions, such as Gaussian noise, downsampling, geometry-based point cloud compression (G-PCC), and video-based point cloud compression (V-PCC).
Table \ref{Tab:existing-dataset} outlines existing publicly available subjective PCQA datasets.

Colored PCQA datasets, such as CPCD 2.0 \citep{Hual21}, SJTU-PCQA \citep{Yangq21}, SIAT-PCQD \citep{wuxj21}, WPC \citep{liuq22}, LS-PCQA \citep{Liuyp22} and BASICS \citep{BASICS2024}, are built for general-purpose PCQA methods.
These datasets contain numerous forms of colored point cloud (i.e., cultural heritages, computer-generated objects, and human figures) that are degraded by diverse geometric and color distortions \citep{Liuyp22}.
Since the color attribute is involved in the subjective quality evaluation experiment, colored PCQA datasets cannot be directly used for geometry-only PCQA methods. 
This is because these datasets are based on MOS ratings, which reflect human visual perception influenced by color. 
As a result, point clouds with the same geometric structure but different color distortions can yield significantly different MOS scores, making these labels unreliable for evaluating geometric differences alone.
By contrast, only a few small-scale colorless PCQA datasets have been constructed for geometry-only PCQA algorithms, such as G-PCD \citep{Alexioue17} and RG-PCD \citep{Alexioue18mex}.
Among them, the largest RG-PCD dataset only contains 6 reference point clouds and 24 geometrically degraded samples.
Therefore, the number of reference point clouds and included distorted samples in existing colorless PCQA datasets are insufficient for the proposed geometry-only ranked-based PCQA algorithm.
Besides, it is also challenging to collect reliable MOS for a large number of point clouds, since the subjective experiment is time-consuming and expensive. 

To alleviate the lack of large-scale subjective datasets, the concept of pseudo MOS is introduced in the IQA \citep{Wujj20,oufz21} and PCQA \citep{Liuyp22}. 
The pseudo MOS, which differs from the MOS rated by subjective experiments, is automatically calculated utilizing selected existing quality assessment algorithms (i.e., using the high-performance full-reference metrics) based on well-designed annotation criteria.
The reliability of the pseudo MOS 
 \citep{Liuyp22} can been verified via conducting various validation experiments.

\subsection{Rank Learning-based Image Quality Assessment}

Rank learning refers to learning a function that can predict the ranking list of a given set of stimuli \citep{Liuty11,Guojf20}, which has been widely used in IQA, since Gao et al. \citep{Gaof15} firstly proposed to apply rank learning to IQA.
Rank learning-based IQA methods aim to rank images instead of assigning absolute quality scores, which can be roughly classified into pairwise-based and list-wise-based methods.

\begin{figure*}[tbp]
	\centering
	\includegraphics[width=0.9\textwidth]{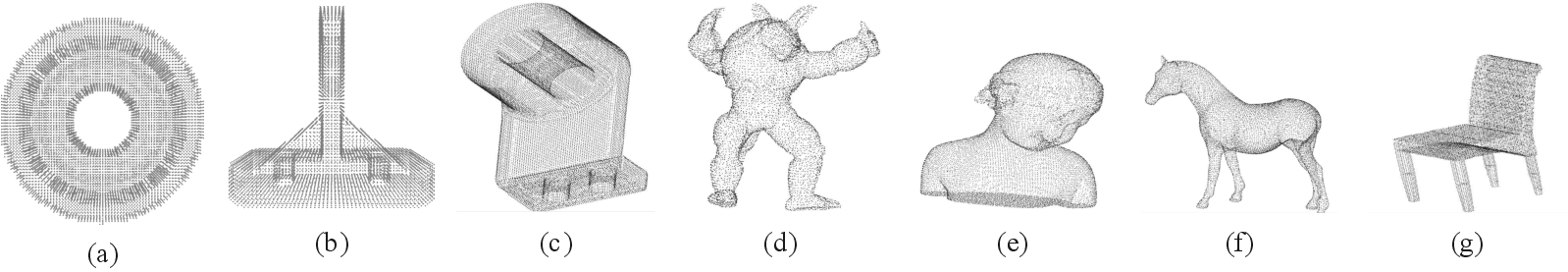}
	\caption{Snapshots of some reference point clouds in the LRL dataset.}
	\label{fig:samples}
\end{figure*}

\textbf{Pairwise-based approaches} assume that the relative quality order between two images can be inferred from other ground-truth formats. 
They focus on optimizing the relative quality preferences between images by minimizing the number of misclassified instance pairs.
Gao et al. \citep{Gaof15} first explored and exploited preference image pairs and utilized natural scene statistics based features to train a robust regression model, in which the preference label represents the relative quality of two images.
In \citep{Mal16}, Ma et al. extracted global image scene textures (GIST) features and resorted to the pairwise rank learning approach to discriminate the perceptual quality between the retargeted image pairs.
Hu et al. \citep{Hub19} proposed to employ DoG-based structure representation, where difference of Gaussian (DoG) is used to decompose the restored image into a set of DoG signals at various octaves and scales, spatial-domain local binary pattern (LBP) feature extraction and frequency-domain log-Gabor feature extraction to extract and concatenate feature vectors to predict the relative quality ranking of restored images.
These methods assess the quality of distorted images via artificially designed unnaturalness expression, which often fail to capture distorted artifacts.
Recently, the siamese network \citep{Liuxl17,Mengxd21,Fuzq22} has been introduced to design deep rank learning-based IQA methods.
They adopt a siamese deep architecture, which takes a pair of degraded images as input and outputs their rank order. 
Each branch is designed to capture deep features of a degraded image using deep neural networks, such as CNN.

\textbf{List-wise-based methods} target constructing loss functions that directly reflect the model's final performance in ranking. 
Gu et al. \citep{Guj19} formulated the no-reference IQA as a recursive list-wise ranking problem that directly learns to rank a list of images with implicit quality measures.
Ou et al. \citep{oufz21} proposed a controllable list-wise ranking loss function to train a CNN by setting an upper and lower bound of rank range and introducing an adaptive margin to tune rank interval. 
They also designed a novel imaging-heuristic approach to generate the rank image samples.
In \citep{Wangxj22}, Wang et al. explored the perceptual quality-related factors and designed a modified list-wise ranking algorithm to achieve a more consistent evaluation with 3D perception and image degradation mechanism in the stereoscopic image re-targeting process.

Inspired by existing list-wise-based IQA methods, this paper reformulates the no-reference GQA as a list-wise ranking problem, which directly learns to rank a list of point clouds with implicit quality measures.

\section{LRL Dataset}
\label{sec:dataset}

To address the dearth of large-scale colorless GQA datasets, a collection of colorless point clouds with diverse content types and geometric complexity is gathered to build a new list-wise ranking  dataset, called LRL, for training the proposed LRL-GQA method.

\subsection{Reference Point Clouds}

In the proposed LRL dataset, a total of 200 colorless point clouds are selected as reference point clouds, which come from existing 3D mesh datasets, such as Stanford 3D scanning repository \citep{TurkG94} and ModelNet \citep{Wuzr15}, as illustrated in Fig. \ref{fig:samples}.
The selected reference point clouds consist of 100 regular-shaped objects and 100 irregular-shaped objects. The regular-shaped objects, examples of which are shown in Fig. \ref{fig:samples} (a), (b), and (c), are industrial parts that are smooth and rectangular, while the irregular-shaped objects, illustrated in Fig. 1(d), (e), (f), and (g), represent real-world objects such as vehicles, people, and daily necessities.
Each selected 3D mesh is firstly normalized to be fully contained in a unit sphere.
Then, a uniformly distributed random sampling strategy is employed to obtain the Cartesian coordinates of the crafted dense point cloud from the surface of mesh. 
Finally, the voxel grid filter is adopted to downsample each point cloud to get a refined sample with regular arrangement and similar density.
The number of points of each reference point cloud ranges between 5K and 50K. 

\begin{figure}[!t]
	\centering
	\includegraphics[width=0.5\textwidth]{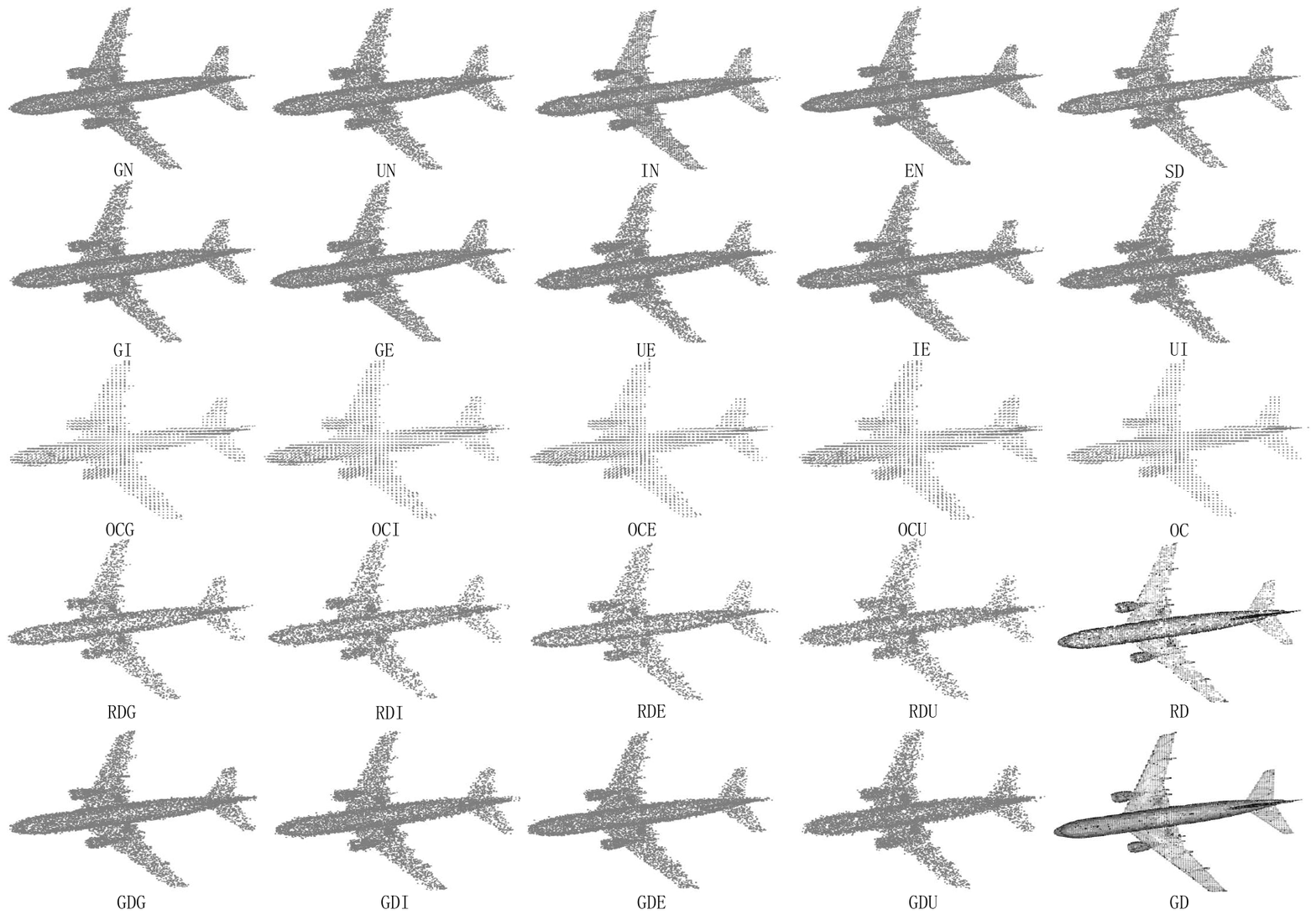}
	\caption{Snapshots of point clouds degraded by different types of distortions.}
	\label{fig:l123}
\end{figure}

\subsection{Distortion Generation}

Each reference point cloud is degraded by 26 types of distortion at 10 different distortion levels, including Gaussian noise (GN), uniform noise (UN), impulse noise (IN), exponential noise (EN), octree-based compression (OC), random downsampling (RD),  grid downsampling (GD), structural distortion (SD) and combination distortions (CD).
Fig. \ref{fig:l123} gives some snapshots of point clouds degraded by different types of distortions. 
For each reference point cloud, the length of the edge between each point and its nearest neighbor is first calculated.
Then, the average length $l_r$  of all edges is employed as the reference value to generate distortions.
The details of each distortion are listed below.

\begin{itemize}
	\item Gaussian noise (GN): Gaussian noise is a kind of noise whose probability density function obeys Gaussian distribution (i.e. normal distribution). The function $normrnd()$ in Matlab is used to add the zero-mean Gaussian noise to point positions with standard deviations of {0.1$l_r$, 0.167$l_r$, 0.233$l_r$, 0.3$l_r$, 0.367$l_r$, 0.433$l_r$, 0.5$l_r$, 0.567$l_r$, 0.633$l_r$, 0.7$l_r$}, respectively.
	
	\item Uniform noise (UN): Uniform noise refers to a type of random variation or interference that has a constant probability distribution over a specified range. The function $rand()$ in Matlab is employed to add the zero-mean uniform noise to point positions through randomly offsetting each point along the $x$, $y$, and $z$ direction independently among [-0.3$l_r$, 0.3$l_r$], [-0.5$l_r$, 0.5$l_r$], [-0.7$l_r$, 0.7$l_r$], [-0.9$l_r$, 0.9$l_r$], [-1.1$l_r$, 1.1$l_r$], [-1.3$l_r$, 1.3$l_r$], [-1.5$l_r$, 1.5$l_r$], [-1.7$l_r$, 1.7$l_r$], [-1.9$l_r$, 1.9$l_r$], and  [-2.1$l_r$, 2.1$l_r$], respectively.
	
	\item Impulse noise (IN): Impulse noise refers to a sudden and short-lived disturbance or interference that occurs in a signal or data. To introduce impulse noise with various distortion levels, we randomly select 10\% points of a point cloud for positive impulse noise and 10\% points for negative impulse noise. The intensity of the impulse noise is set to 0.3$l_r$, 0.5$l_r$, 0.7$l_r$, 0.9$l_r$, 1.1$l_r$, 1.3$l_r$, 1.5$l_r$, 1.7$l_r$, 1.9$l_r$, and 2.1$l_r$, respectively.
	
	\item Exponential noise (EN): Exponential noise refers to a type of random variation or disturbance that follows an exponential probability distribution. The function $exprnd()$ in Matlab is adopted to add exponential noise to each point along the $x$, $y$, and $z$ direction independently with the mean parameter 0.1$l_r$, 0.167$l_r$, 0.233$l_r$, 0.3$l_r$, 0.367$l_r$, 0.433$l_r$, 0.5$l_r$, 0.567$l_r$, 0.633$l_r$, and 0.7$l_r$, respectively.
	
	\item Octree-based compression (OC): Octree-based compression is a data compression technique that utilizes octrees to compress 3D volumetric data, such as 3D models or point clouds. The function   provided by the well-known Point Cloud Library (i.e. $OctreePointCloudCompression()$) is used to compress each point cloud by setting the octree resolution at 0.01$l_r$, 0.0117$l_r$, 0.0133$l_r$, 0.015$l_r$, 0.0167$l_r$, 0.0183$l_r$, 0.02$l_r$, 0.0217$l_r$, 0.0233$l_r$, and 0.025$l_r$ respectively.
	
	\item Random downsampling (RD):  Random downsampling is a technique used to reduce the size of a point cloud by randomly selecting a subset of the original points. The function $pcdownsampling()$  in Matlab is employed to randomly downsample the point cloud by removing 15\%, 21.1\%, 27.2\%, 33.3\%, 39.4\%, 45.6\%, 51.7\%, 57.8\%, 63.9\%, and 70\% points from the original point, respectively.
	
	\item Grid downsampling (GD): Grid downsampling is a technique used to reduce the resolution or density of a point cloud by systematically selecting points based on a regular grid pattern. The function $pcdownsampling()$ in Matlab is used to downsample the point cloud through a sampling grid with the resolution 1.2$l_r$, 1.34$l_r$, 1.49$l_r$, 1.63$l_r$, 1.78$l_r$, 1.92$l_r$, 2.06$l_r$, 2.21$l_r$, 2.36$l_r$, and 2.5$l_r$, respectively.
 
    \item Structural distortion (SD): Structural distortion comes from the time-of-flight camera in the BlenSor software \citep{BlenSor}. Different sampling levels are realized by adjusting the opening angle of the camera among 25°, 30°, 35°, 40°, 45°, 50°, 55°, 60°, 65°, and 70°, respectively. The smaller the opening angle, the lower the distortion of obtained point clouds.

    \item Combination distortions (CD): The combination of four varieties of noises (GN, UN, IN and EN) yields six composite distortions, which are utilized to emulate the hybrid distortions that might emerge in the acquisition process. These are specifically GN\&UN (GU), GN\&IN (GI), GN\&EN (GE), UN\&IN (UI), UN\&EN (UE), and IN\&EN (IE). Meanwhile, the 12 combinations derived from four types representing  noises (GN, UN, IN and EN) in acquisition and three types of compression distortions (OC, RD and GD) in transmission are individually named as OC\&GN (OCG), OC\&UN (OCU), OC\&IN (OCI), OC\&EN (OCE), RD\&GN (RDG), RD\&UN (RDU), RD\&IN (RDI), RD\&EN (RDE), GD\&GN (GDG), GD\&UN (GDU), GD\&IN (GDI), and GD\&EN (GDE).

\end{itemize}

Therefore, each reference point cloud has $26 \times 10$ degraded samples.
In total, there are $200 + 200 \times 26 \times 10 = 52200$ point clouds with a wide range of visual quality levels in the LRL dataset.

\section{Proposed Method}
\label{sec:Proposed Method}

\begin{figure*}[!tbp]
	\centering
	\includegraphics[width=0.75\textwidth]{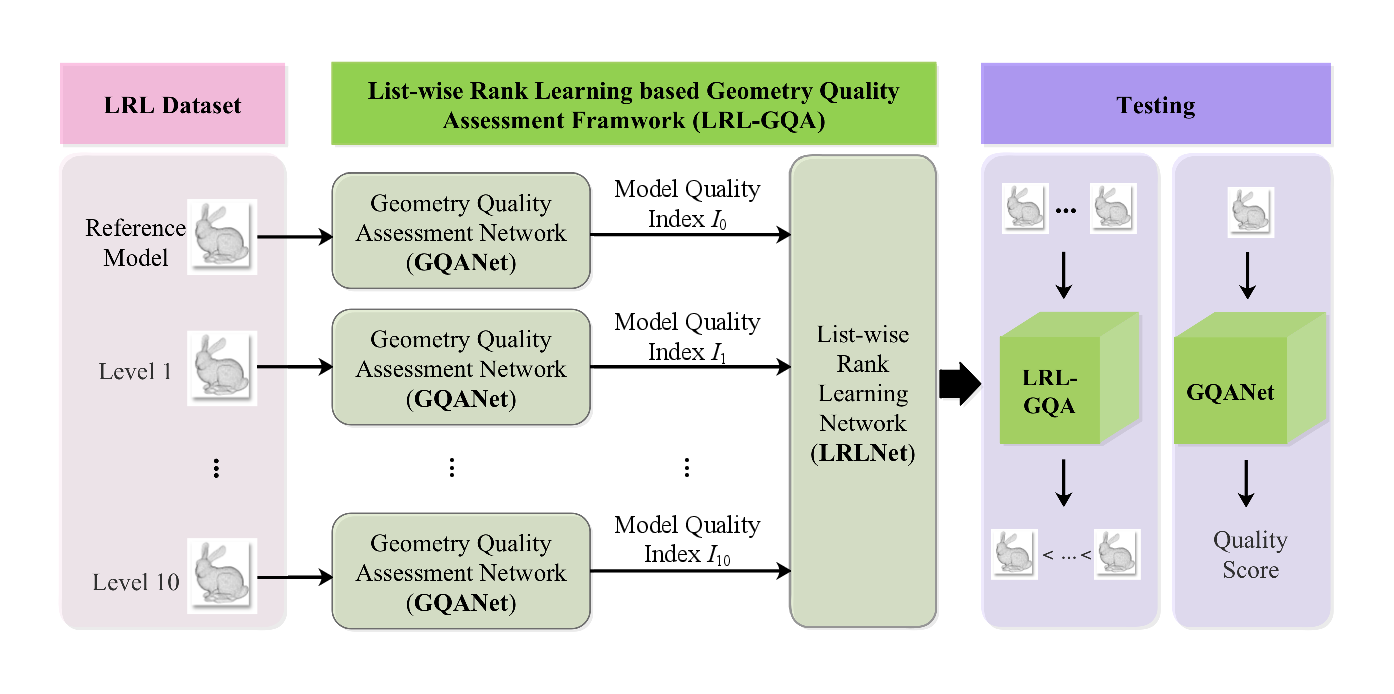}
	\caption{Overview of the proposed LRL-GQA framework. The LRL dataset contains ranked distorted point clouds for training and testing. The LRL-GQA consists of a geometry quality assessment network (GQANet) and a list-wise rank learning network (LRLNet). Based on the LRL dataset, the LRL-GQA can be trained to rank the list of degraded point clouds, as well as calculate absolute quality scores after fine-tuning.}
	\label{fig:Overview}
\end{figure*}

The proposed LRL-GQA method aims to learn the global ranking of distorted point clouds using list-wise rank learning from the entire list.
It first encodes a list of distorted point clouds into patch-wise geometric features using the GQANet. Subsequently, the LRLNet is applied to these features to obtain list-wise ranking results, as illustrated in Fig. \ref{fig:Overview}.

\begin{figure}[!t]
	\centering
	\includegraphics[width=0.45\textwidth]{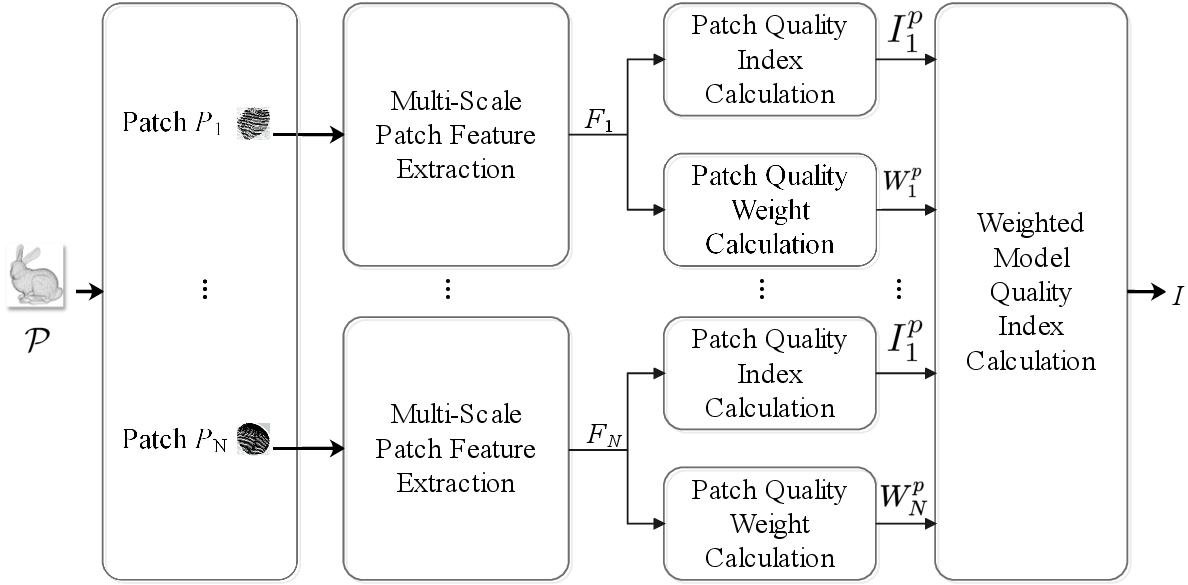}
	\caption{Overview of the GQANet framework. Given an input point cloud $\mathcal{P}$, the GQANet outputs its model quality index $I$.}
	\label{fig:GQA}
\end{figure}

\subsection{Geometry Quality Assessment Network}
The GQANet is designed to capture multi-scale features of point clouds at the patch level and ultimately aggregates these features by weighting the patches according to their significance within the entire point cloud.
Specifically, it comprises patch generation,  multi-scale patch feature extraction (MPFE), patch quality index calculation, patch quality weight calculation, and weighted model quality index calculation, as shown in Fig. \ref{fig:GQA}.
Given an input point cloud $\mathcal{P}$, the GQANet outputs its quality index $I$ which is defined in Eq.~\eqref{eq:f}:
\begin{equation}
	\label{eq:f}
	I  = f (\mathcal{P}).
\end{equation}
First, the patch generation segments each input point cloud $\mathcal{P}$ into different overlapping patches. 
Then, for each patch, the MPFE uses edge convolution to extract patch-wise multi-scale quality-aware features, which is then used to calculate the patch quality index and patch quality weight.
Finally, a weighted model quality index calculation module is designed to predict the final model quality index $I$ of $\mathcal{P}$.

\begin{figure}[!t]
	\centering
	\includegraphics[width=0.45\textwidth]{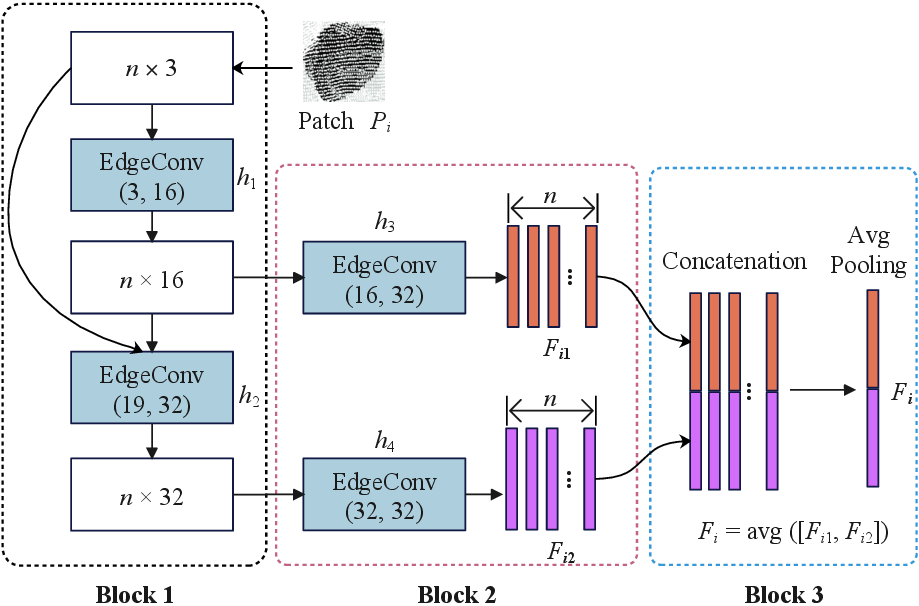}
	\caption{Structure of the multi-scale patch feature extraction (MPFE) module in the GQANet. For each patch $P_i$, the MPFE module employs edge convolution to extract its multi-scale quality-aware features $F_i$.}
	\label{fig:Feature_extracttion}
\end{figure}

\subsubsection{Patch Generation}

The input point cloud is segmented into $N$ different overlapping patches, considering that those salient regions (e.g., the region with high curvature) in the point cloud should play a more important role in the quality assessment than other plain regions \citep{Hual21}.
Different weights will be assigned to these patches to describe their impact on the final quality evaluation

First, for each point cloud $\mathcal{P}$, the farthest point sampling (FPS) strategy \citep{qi2017pointnet++} is employed to uniformly sample $N$ anchor points.
Undoubtedly, for each reference point cloud, the quality ranking among patches of its different degraded samples should be performed on the same region.
However, the degraded samples of the same reference point cloud may have different number of points since some distortions, such as downsampling, may reduce the number of points.
To address this issue, these sampled anchor points can be used to generate patches for all degraded samples of their reference point cloud.
Specifically, for each anchor point $p_{i}$, the ball query with its radius $r$ is used to find $n$ neighborhoods to form the patch $P_i$. 
Formally, the patch $P_i$ centered at $p_{i}$ is defined in Eq. (\ref{eq:patch}):
\begin{equation}
    \label{eq:patch}
    P_i = \left\{ p_j \mid \left\| p_j - p_i \right\| < r \right\} \in \mathbb{R}^{n \times 3}.
\end{equation}
To effectively tune network parameters with batches, each patch should contain the same number of points.
This paper pads the $p_{i}$ for patches with insufficient points ( $<  n$)  and do random down-sampling for patches with sufficient points ( $>  n$).

\subsubsection{Multi-Scale Patch Feature Extraction}

The MPFE module consists of three blocks to extract multi-scale quality perception features $F_{i}$ of each patch $P_{i}$ defined in Eq. (\ref{eq:patch}), as depicted in Fig. \ref{fig:Feature_extracttion}. 

The first block consists of two layers of edge convolution (EdgeConv) operations \citep{Wangy19}, which are used to extract different scale features through Eq. (\ref{eq:multi-scale}):
\begin{equation}
        \label{eq:multi-scale}
	\begin{split} 
		\left\{\begin{matrix} 
			F_{i1} = &h_{3}(h_{1}(P_{i})) \\
			F_{i2} = &h_{4}([h_{1}(P_{i}),h_{2}(h_{1}(P_{i}))]) 
		\end{matrix}\right. ,
	\end{split}
\end{equation}
where $F_{i1}\in \mathbb{R}^{32} $, $F_{i2}\in \mathbb{R}^{32} $, $ [\cdot] $ indicates the feature concatenation, $ h_{1}$ , $ h_{2}$ , $ h_{3}$ , and $ h_{4} $ are the full EdgeConv layers, respectively.
The second block has two branches, each of which is composed of a layer of EdgeConv, aiming to adjust features of different scales to an appropriate dimension. 
The third block concatenates the different scale features $F_{i1}$ and $F_{i2}$, and performs average pooling to aggregate the multi-scale features $F_{i}$ ($F_{i}\in \mathbb{R}^{64} $) through Eq. (\ref{eq:avg-pooling}):

\begin{equation}
    \label{eq:avg-pooling}
	F_{i} = {\rm avg}([F_{i1}, F_{i2}]).
\end{equation}

\subsubsection{Patch Quality Index Calculation}
\label{subsubsec:Patch Score}

\begin{figure}[!t]
	\centering
	\includegraphics[width=0.5\textwidth]{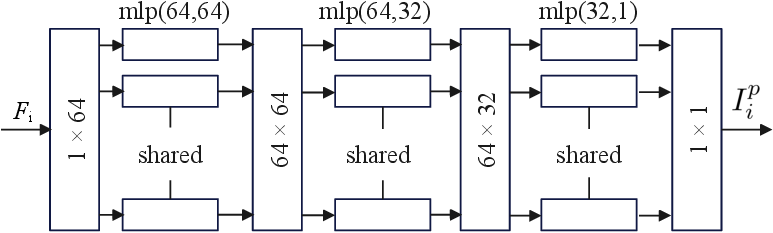}
	\caption{Structure of the patch quality index calculation module in the GQANet. Given the feature vector $F_i$ of each patch , this module predicts its patch quality index $I^p_i$.}
	\label{fig:patch_score}
\end{figure}

The patch quality index calculation module consists of three multi-layer perceptions (MLPs),  as shown in Fig \ref{fig:patch_score}.
The module predicts the quality index $I_{i}^{p}$ to represent the geometric quality of $P_i$ for a given patch $P_i$ with the multi-scale feature $F_i$ defined in Eq. (\ref{eq:avg-pooling}).

\subsubsection{Patch Quality Weight Calculation}
\label{subsubsec:Patch Weight}

\begin{figure}[!t]
	\centering
	\includegraphics[width=0.5\textwidth]{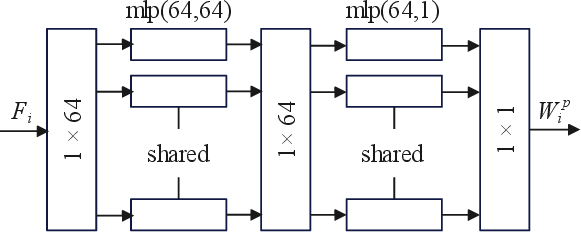}
	\caption{Structure of the patch quality weight calculation in the GQANet. Given the feature vector $F_i$ of each patch, this module predicts its weight $W_{i}^{p}$.}
	\label{fig:model_score}
\end{figure}

The patch quality weight calculation module predicts the weight $W_{i}^{p}$ of each patch, considering that patches with fine geometric details should be assigned more weights in the pooling phase.
Specifically, the module is composed of two MLPs, as illustrated in Fig. \ref{fig:model_score}.

\subsubsection{Weighted Model Quality Index Calculation}
\label{subsubsec:Model Score}

Given the quality index $I_{i}^{p}$ and weight $W_{i}^{p}$ of each patch, the weighted model quality index calculation module computes the quality index $I$ of the point cloud $\mathcal{P}$ through Eq. (\ref{eq:quality-index}):
\begin{equation}
    \label{eq:quality-index}
	I = \frac{\sum_{i=1}^{N}W_{i}^{p}I_{i}^{p} }{\sum_{i=1}^{N}W_{i}^{p}}.
\end{equation}

\subsection{List-wise Rank Learning}
The LRLNet predicts the relative quality of all the point clouds associated with the same group according to their predicted quality indexes. 
Inspired by the list-wise ranking method proposed by Xia \citep{Xiaf08}, the LRLNet deploys the listMLE to globally optimize the patch quality index calculation and weighted model quality index calculation modules, compared to the pair-wise approaches \citep{Liuty11,Wangxj22}.

Given that $\{\mathcal{P}_d^i \mid i=0,i,\cdots, 10\}$ is a mini-batch, which consists of a pristine point cloud $\mathcal{P}_d^0$ and its 10 degraded versions $\mathcal{P}_d^k$ $ (k=1,\cdots, 10)$ at different degradation levels yet the same type of distortion.
Let $y_i$ $(y_i \in[0,10], i=0,1,\cdots, 10)$ be the index of $\mathcal{P}_d^i$ which is ranked at position $i$.
All the point clouds $\mathcal{P}_d^i$ in the mini-batch are sorted by their ground-truth ranking label $y_i$, rather than their predicted quality index $I_d^i = f (\mathcal{P}_d^i)$ defined in Eq. (\ref{eq:quality-index}). 
The ranking function $L$ in Eq.~\eqref{eq:ranking-loss} is defined as a function over the ground-truth list of point clouds sorted by $y_i$ and the list of point clouds sorted by $f(\mathcal{P}_d)$,
\begin{equation}
    \label{eq:ranking-loss}
	L(f(\mathcal{P}_d), y) = - \log (P(y \mid \mathcal{P}_d; f)),
\end{equation}
where
\begin{equation}
    \label{eq:probability-function}
	P(y \mid \mathcal{P}_d; f)=\prod\limits_{i=0}^{10}\frac{\exp(f(\mathcal{P}_d^{y_i}))}{\sum_{k=i}^{10}\exp(f(\mathcal{P}_d^{y_k}))},
\end{equation}
where $f(\cdot)$ denotes the model quality index calculation function described in Eq. (\ref{eq:f}).
	
\subsection{Training and Fine-tuning}
\label{subsec:Training-tuning}

\begin{table*}[!t]
\setlength{\tabcolsep}{4pt}
\centering
\caption{NDCG scores of methods on the LRL dataset with 10 distortion levels. Higher values imply better outcomes.}
\label{tab:NDCG10}
\resizebox{\textwidth}{!}{
\begin{tabular}{cccccccccccccccc}
\hline
\multicolumn{1}{c|}{}                    & \multicolumn{1}{c|}{Types}                                                & GN                   & IN                   & EN                   & UN                   & OC                   & RD                   & GD                   & SD                   & GU                   & GI                   & GE                   & UI                                  & UE                   & IE                   \\ \hline
\multicolumn{1}{c|}{\multirow{11}{*}{FR}} & \multicolumn{1}{c|}{Po2Po$_{MSE}$\citep{Rufaelm16}}       & \textbf{1.000}       & \textbf{1.000}       & \textbf{1.000}       & \textbf{1.000}       & 0.998                & \textbf{1.000}       & \textbf{1.000}       & \textbf{1.000}       & \textbf{1.000}       & \textbf{1.000}       & \textbf{1.000}       & \textbf{1.000}                      & \textbf{1.000}       & \textbf{1.000}       \\
\multicolumn{1}{c|}{}                    & \multicolumn{1}{c|}{Po2Po$_{HD}$\citep{Rufaelm16}}       & 1.000                & 1.000                & 0.999                & 1.000                & \textbf{1.000}       & 0.999                & 1.000                & 0.993                & 1.000                & 1.000                & 0.999                & 1.000                               & 1.000                & 1.000                \\
\multicolumn{1}{c|}{}                    & \multicolumn{1}{c|}{Po2Po$_{PSNR}$\citep{Rufaelm16}}      & 1.000                & 1.000                & 1.000                & 1.000                & 0.998                & 1.000                & 1.000                & 1.000                & 1.000                & 1.000                & 1.000                & 1.000                               & 1.000                & 1.000                \\
\multicolumn{1}{c|}{}                    & \multicolumn{1}{c|}{Po2Pl$_{MSE}$\citep{Tiand17}}         & 1.000                & 1.000                & 1.000                & 1.000                & 0.996                & 1.000                & 1.000                & 1.000                & 1.000                & 1.000                & 1.000                & 1.000                               & 1.000                & 1.000                \\
\multicolumn{1}{c|}{}                    & \multicolumn{1}{c|}{Po2Pl$_{HD}$\citep{Tiand17}}         & 1.000                & 1.000                & 1.000                & 1.000                & 1.000                & 1.000                & 0.999                & 0.993                & 1.000                & 1.000                & 0.999                & 1.000                               & 0.999                & 1.000                \\
\multicolumn{1}{c|}{}                    & \multicolumn{1}{c|}{Po2Pl$_{PSNR} $\citep{Tiand17}}       & 1.000                & 1.000                & 1.000                & 1.000                & 0.998                & 1.000                & 0.995                & 1.000                & 1.000                & 1.000                & 1.000                & 1.000                               & 1.000                & 1.000                \\
\multicolumn{1}{c|}{}                    & \multicolumn{1}{c|}{Pl2Pl$_{MSE} $\citep{Alexioue18icme}} & 0.999                & 0.996                & 1.000                & 0.998                & 1.000                & 0.929                & 0.929                & 1.000                & 1.000                & 0.996                & 1.000                & 0.999                               & 0.998                & 0.999                \\
\multicolumn{1}{c|}{}                    & \multicolumn{1}{c|}{Pl2Pl$_{HD}$\citep{Alexioue18icme}}  & 0.972                & 0.880                & 0.975                & 0.990                & 0.850                & 0.929                & 0.929                & 0.850                & 0.991                & 0.982                & 0.990                & 0.985                               & 0.988                & 0.993                \\
\multicolumn{1}{c|}{}                    & \multicolumn{1}{c|}{Pl2Pl$_{PSNR}$\citep{Alexioue18icme}} & 0.992                & 0.959                & 0.994                & 0.990                & 0.992                & 0.850                & 0.850                & 1.000                & 0.991                & 0.977                & 0.988                & 0.998                               & 0.975                & 0.990                \\
\multicolumn{1}{c|}{}                    & \multicolumn{1}{c|}{PCQM \citep{Meynetg20}}                & 0.997 &0.960 &0.999 &1.000 &0.921 &0.885 &0.955 &1.000 &1.000 &1.000 &1.000 &1.000 &1.000 &1.000             \\
\multicolumn{1}{c|}{}                    & \multicolumn{1}{c|}{PointSSIM \citep{Alexioue20}}                                            & 1.000                & 0.968                & 1.000                & 1.000                & 1.000                & 1.000                & 0.996                & 0.999                & 1.000                & 1.000                & 1.000                & 1.000                               & 1.000                & 1.000                \\ \hline
\multicolumn{1}{c|}{\multirow{2}{*}{NR}}             & \multicolumn{1}{c|}{3DTA \citep{zhu20243dta}}                                                  & 0.895 &0.881 &0.992 &0.928 &0.914 &0.934 &0.999 &0.894 &0.994 &0.932 &0.892 &0.969 &0.861 &0.981 \\
\multicolumn{1}{c|}{}                    & \multicolumn{1}{c|}{ours}                                                 & 0.993                & 0.994                & 0.997                & 0.996                & 0.996                & 0.994                & 0.995                & 0.993                & 0.992                & 0.997                & 0.994                & 0.995                               & 0.994                & 0.993                \\ \hline
                                         &                                                                           &                      &                      &                      &                      &                      &                      &                      &                      &                      &                      &                      &                                     &                      &                      \\ \hline
\multicolumn{1}{c|}{}                    & \multicolumn{1}{c|}{Types}                                                & OCG                  & OCU                  & OCI                  & OCE                  & RDG                  & RDU                  & RDI                  & RDE                  & GDG                  & GDU                  & GDI                  & \multicolumn{1}{c|}{GDE}            & \multicolumn{2}{c}{MEAN}                    \\ \hline
\multicolumn{1}{c|}{\multirow{11}{*}{FR}} & \multicolumn{1}{c|}{Po2Po$_{MSE}$\citep{Rufaelm16}}       & 0.998                & \textbf{1.000}       & \textbf{1.000}       & \textbf{1.000}       & \textbf{1.000}       & \textbf{1.000}       & \textbf{1.000}       & \textbf{1.000}       & \textbf{1.000}       & \textbf{1.000}       & \textbf{1.000}       & \multicolumn{1}{c|}{\textbf{1.000}} & \multicolumn{2}{c}{\textbf{1.000}}          \\
\multicolumn{1}{c|}{}                    & \multicolumn{1}{c|}{Po2Po$_{HD}$\citep{Rufaelm16}}       & \textbf{1.000}       & 1.000                & 1.000                & 0.999                & 1.000                & 1.000                & 1.000                & 1.000                & 1.000                & 1.000                & 1.000                & \multicolumn{1}{c|}{1.000}          & \multicolumn{2}{c}{1.000}                   \\
\multicolumn{1}{c|}{}                    & \multicolumn{1}{c|}{Po2Po$_{PSNR}$\citep{Rufaelm16}}      & 1.000                & 1.000                & 1.000                & 1.000                & 1.000                & 1.000                & 1.000                & 1.000                & 1.000                & 1.000                & 1.000                & \multicolumn{1}{c|}{1.000}          & \multicolumn{2}{c}{1.000}                   \\
\multicolumn{1}{c|}{}                    & \multicolumn{1}{c|}{Po2Pl$_{MSE}$\citep{Tiand17}}         & 0.998                & 1.000                & 0.998                & 1.000                & 1.000                & 1.000                & 1.000                & 1.000                & 1.000                & 1.000                & 1.000                & \multicolumn{1}{c|}{1.000}          & \multicolumn{2}{c}{1.000}                   \\
\multicolumn{1}{c|}{}                    & \multicolumn{1}{c|}{Po2Pl$_{HD}$\citep{Tiand17}}         & 1.000                & 1.000                & 1.000                & 1.000                & 1.000                & 1.000                & 1.000                & 1.000                & 1.000                & 1.000                & 1.000                & \multicolumn{1}{c|}{1.000}          & \multicolumn{2}{c}{1.000}                   \\
\multicolumn{1}{c|}{}                    & \multicolumn{1}{c|}{Po2Pl$_{PSNR} $\citep{Tiand17}}       & 1.000                & 1.000                & 1.000                & 1.000                & 1.000                & 1.000                & 1.000                & 1.000                & 1.000                & 1.000                & 1.000                & \multicolumn{1}{c|}{1.000}          & \multicolumn{2}{c}{1.000}                   \\
\multicolumn{1}{c|}{}                    & \multicolumn{1}{c|}{Pl2Pl$_{MSE} $\citep{Alexioue18icme}} & 0.997                & 0.999                & 0.997                & 0.994                & 1.000                & 1.000                & 1.000                & 1.000                & 1.000                & 0.999                & 0.999                & \multicolumn{1}{c|}{0.999}          & \multicolumn{2}{c}{0.993}                   \\
\multicolumn{1}{c|}{}                    & \multicolumn{1}{c|}{Pl2Pl$_{HD}$\citep{Alexioue18icme}}  & 0.878                & 0.879                & 0.874                & 0.875                & 0.996                & 0.995                & 0.995                & 0.997                & 0.986                & 0.994                & 0.990                & \multicolumn{1}{c|}{0.988}          & \multicolumn{2}{c}{0.952}                   \\
\multicolumn{1}{c|}{}                    & \multicolumn{1}{c|}{Pl2Pl$_{PSNR}$\citep{Alexioue18icme}} & 0.971                & 0.974                & 0.993                & 0.998                & 1.000                & 0.999                & 1.000                & 1.000                & 0.998                & 0.987                & 0.977                & \multicolumn{1}{c|}{0.994}          & \multicolumn{2}{c}{0.978}                   \\
\multicolumn{1}{c|}{}                    & \multicolumn{1}{c|}{PCQM \citep{Meynetg20}}                                                 & 0.997 &0.999 &0.920 &0.921 &0.999 &1.000 &0.992 &1.000 &0.999 &1.000 &0.959& \multicolumn{1}{c|}{0.991} & \multicolumn{2}{c}{0.981}              \\
\multicolumn{1}{c|}{}                    & \multicolumn{1}{c|}{PointSSIM \citep{Alexioue20}}                                            & 0.997                & 0.999                & 1.000                & 0.999                & 1.000                & 1.000                & 1.000                & 1.000                & 1.000                & 1.000                & 1.000                & \multicolumn{1}{c|}{1.000}          & \multicolumn{2}{c}{0.998}                   \\ \hline
\multicolumn{1}{c|}{\multirow{2}{*}{NR}}      & \multicolumn{1}{c|}{3DTA \citep{zhu20243dta}}                                                  & 0.953 &0.968 &0.890 &0.884 &0.947 &0.889 &0.904 &0.912 &0.916 &0.986 &0.888  & \multicolumn{1}{c|}{0.983} & \multicolumn{2}{c}{0.930}            \\

\multicolumn{1}{c|}{}                    & \multicolumn{1}{c|}{ours}                                                 & 0.993                & 0.993                & 0.992                & 0.994                & 0.995                & 0.994                & 0.993                & 0.994                & 0.994                & 0.993                & 0.995                & \multicolumn{1}{c|}{0.992}          & \multicolumn{2}{c}{0.994}                   \\ \hline
\end{tabular}}
\end{table*}

The end-to-end training of the entire LRL-GQA network is a challenging task and it is also difficult to converge to achieve higher accuracy for point clouds.
Therefore, to facilitate the training process, the MPFE module is initially pre-trained through a distortion-level classification task.
After that, the entire LRL-GQA network is trained in an end-to-end way with the MPFE module's parameters frozen on the LRL dataset.
Finally, in addition to predicting the relative geometric quality of input point clouds, the proposed GQANet is capable of calculating absolute subjective scores after fine-tuning. 
Due to the randomness of the adopted FPS strategy in the patch generation module, it should be pointed out that the generated patches of each point cloud remain fixed throughout the training stage.
Only in this way can the LRL-GQA network be properly trained.

\subsubsection{Pre-Training of the MPFE Module} 

The MPFE module is pre-trained on the LRL dataset by introducing a patch-wise distortion-level classification task.
Specifically, the output of the MPFE module is fed into a MLP ($64 \times 64 \times 32 \times 11$) to classify the distortion-level of each patch.
To that end, all reference point clouds from the LRL dataset are randomly divided into two non-overlapping subsets: 90\% for training and 10\% for testing.
There are 1 reference category and 10 distortion categories according to the construction of the LRL dataset, resulting in 11 distortion-level categories in total.
During the training stage, the batch size is set to 1.
In other words, for each input point cloud, the batch actually contains $N$ patches associated with the same distortion type.
The MPFE module is trained by using the cross entropy loss function to achieve quality-aware features and fine distortion level recognition ability.

\subsubsection{Fine-tuning for Quality Score Calculation } 

The pre-trained GQANet can be fine-tuned to calculate absolute subjective quality scores. 
For a batch of $K$ point clouds with subjective scores, the fine-tuning process minimizes the squared Euclidean distance between the predicted scores $\hat{S}_{i}$ and the ground-truth scores $S_{i}$ through Eq. (\ref{eq:mse-loss}):
\begin{equation}
    \label{eq:mse-loss}
	L\left(S_{i},\hat{S}_{i}\right)=\frac{1}{K}\sum_{i=1}^{K}\left(S_{i}-\hat{S}_{i}\right)^{2} ,
\end{equation}
where $ S_{i}$ is the subjective score of the $i-$th point cloud, and $ \hat{S}_{i}$ is the prediction score of the $i-$th point cloud.
Before fine-tuning, the pre-trained GQANet just outputs relative quality indexes.
After fine-tuning, the fine-tuned GQANet is able to predict absolute subjective scores.
Note that this paper does not perform any fine-tuning on the MPFE module whereby its parameters would have been modified under the supervision of ground-truth quality scores. 
However, as shown in Sec. \ref{subsec:score}, compared with existing full-reference GQA metrics, the GQANet yields competitive performance on the LRL-PMOS dataset, even without fine-tuning of the MPFE module.

\section{Experimental Results and Analysis}
\label{sec:Experimental}

\begin{figure*}[!h]
	\centering
	\includegraphics[width=0.9\textwidth]{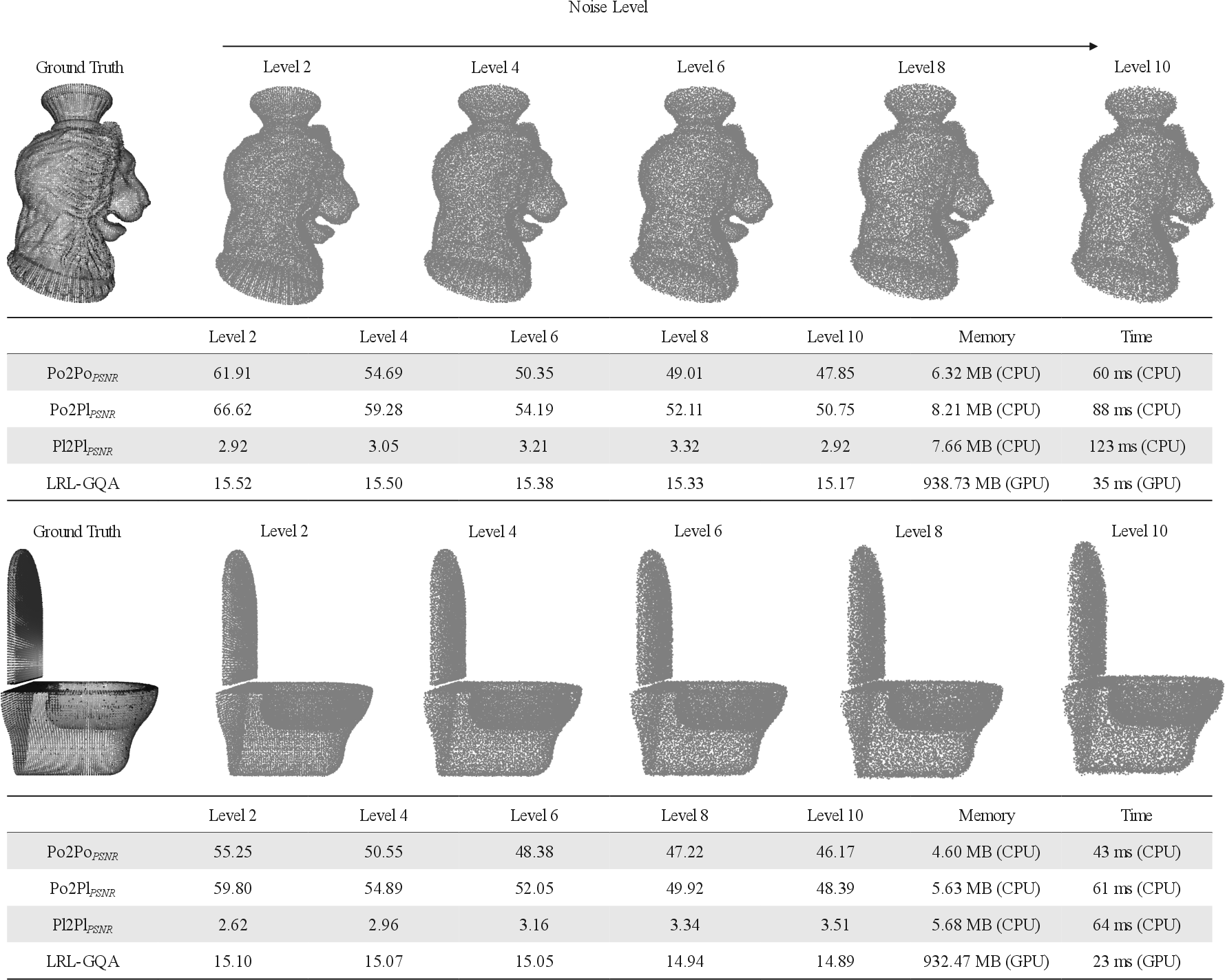}
	\caption{Visual comparison of point cloud quality scores of compared methods. The memory usage and computation time of each method are also shown in the figure.}
	\label{fig:existing_techniques_visually}
\end{figure*}

\subsection{Experimental settings}

\subsubsection{Parameters Selection}

For the MPFE module, each point cloud is empirically segmented into $N = 64$ overlapping patches with the patch radius $r = 0.2$.
Each patch contains $n = 512$ points.
To pre-train the MPFE module via the auxiliary classification task, the Leaky ReLU is used as the activation function. 
And, the cross entropy loss function and Adam optimizer are also employed.
The initial learning rate is set to 0.001,  and the training epochs is set to 240.
For the LRL module, the Adam optimizer is used for training.
The initial learning rate is set to 0.0001.
The number of epochs is set to 800.
Considering the randomness of patch generation, this paper conducts the experiments for 4 times and records the average value for each experimental setting to ensure the reliability of the experimental results.

\subsubsection{Compared Methods}

\begin{table*}[!t]
\setlength{\tabcolsep}{4pt}
\centering
\caption{NDCG scores of different methods on the LRL dataset with 20 distortion levels. Higher values imply better outcomes.}
        \label{tab:NDCG20}
\resizebox{0.95\textwidth}{!}{
\begin{tabular}{cccccccccccccccc}
\hline
\multicolumn{1}{c|}{}                     & \multicolumn{1}{c|}{Types}                                                & GN             & IN             & EN             & UN             & OC             & RD             & GD             & SD             & GU             & GI             & GE             & UI                                  & UE               & IE              \\ \hline
\multicolumn{1}{c|}{\multirow{11}{*}{FR}} & \multicolumn{1}{c|}{Po2Po$_{MSE}$\citep{Rufaelm16}}       & \textbf{1.000} & \textbf{1.000} & \textbf{1.000} & \textbf{1.000} & 0.999          & \textbf{1.000} & \textbf{1.000} & \textbf{1.000} & \textbf{1.000} & \textbf{1.000} & \textbf{1.000} & \textbf{1.000}                      & \textbf{1.000}   & \textbf{1.000}  \\
\multicolumn{1}{c|}{}                     & \multicolumn{1}{c|}{Po2Po$_{HD}$\citep{Rufaelm16}}       & 1.000          & 1.000          & 1.000          & 1.000          & \textbf{1.000} & 0.999          & 0.999          & 0.994          & 1.000          & 1.000          & 1.000          & 1.000                               & 1.000            & 1.000           \\
\multicolumn{1}{c|}{}                     & \multicolumn{1}{c|}{Po2Po$_{PSNR}$\citep{Rufaelm16}}      & 1.000          & 1.000          & 1.000          & 1.000          & 0.999          & 1.000          & 1.000          & 1.000          & 1.000          & 1.000          & 1.000          & 1.000                               & 1.000            & 1.000           \\
\multicolumn{1}{c|}{}                     & \multicolumn{1}{c|}{Po2Pl$_{MSE}$\citep{Tiand17}}         & 1.000          & 1.000          & 1.000          & 1.000          & 0.997          & 1.000          & 1.000          & 1.000          & 1.000          & 1.000          & 1.000          & 1.000                               & 1.000            & 1.000           \\
\multicolumn{1}{c|}{}                     & \multicolumn{1}{c|}{Po2Pl$_{HD}$\citep{Tiand17}}         & 1.000          & 1.000          & 1.000          & 1.000          & 1.000          & 0.998          & 1.000          & 0.994          & 1.000          & 1.000          & 1.000          & 1.000                               & 1.000            & 0.999           \\
\multicolumn{1}{c|}{}                     & \multicolumn{1}{c|}{Po2Pl$_{PSNR} $\citep{Tiand17}}       & 1.000          & 1.000          & 1.000          & 1.000          & 1.000          & 1.000          & 1.000          & 1.000          & 1.000          & 1.000          & 1.000          & 1.000                               & 1.000            & 1.000           \\
\multicolumn{1}{c|}{}                     & \multicolumn{1}{c|}{Pl2Pl$_{MSE} $\citep{Alexioue18icme}} & 0.999          & 0.994          & 1.000          & 0.998          & 0.999          & 0.922          & 0.925          & 0.998          & 0.999          & 0.994          & 0.998          & 0.996                               & 0.994            & 0.999           \\
\multicolumn{1}{c|}{}                     & \multicolumn{1}{c|}{Pl2Pl$_{HD}$\citep{Alexioue18icme}}  & 0.934          & 0.889          & 0.967          & 0.990          & 0.998          & 0.915          & 0.922          & 0.849          & 0.986          & 0.982          & 0.992          & 0.979                               & 0.988            & 0.992           \\
\multicolumn{1}{c|}{}                     & \multicolumn{1}{c|}{Pl2Pl$_{PSNR}$\citep{Alexioue18icme}} & 0.990          & 0.956          & 0.993          & 0.989          & 0.991          & 0.861          & 0.849          & 0.992          & 0.990          & 0.968          & 0.984          & 0.996                               & 0.972            & 0.990           \\
\multicolumn{1}{c|}{}                     & \multicolumn{1}{c|}{PCQM \citep{Meynetg20}}                                                 & 0.963          & 0.915          & 0.923          & 0.903          & 0.916          & 0.978          & 0.911          & 0.936          & 0.889          & 0.945          & 0.903          & 0.960                               & 0.927            & 0.957           \\
\multicolumn{1}{c|}{}                     & \multicolumn{1}{c|}{PointSSIM \citep{Alexioue20}}                                            & 1.000          & 1.000          & 1.000          & 1.000          & 1.000          & 1.000          & 1.000          & 1.000          & 1.000          & 0.939          & 1.000          & 1.000                               & 1.000            & 0.997           \\ \hline
\multicolumn{1}{c|}{\multirow{2}{*}{NR}}      & \multicolumn{1}{c|}{3DTA \citep{zhu20243dta}}                                                    & 0.920 &0.995 &0.929 &0.896 &0.937 &0.989 &0.890 &0.944 &0.911 &0.892 &0.984 &0.916 &0.872 &0.907 \\

\multicolumn{1}{c|}{}                    & \multicolumn{1}{c|}{ours}                                                 & 0.994  & 0.993  & 0.994  & 0.992  & 0.995  & 0.993  & 0.991  & 0.995  & 0.994  & 0.994  & 0.992  & 0.991  & 0.994  & 0.995               \\ \hline
                                          &                                                                           &                &                &                &                &                &                &                &                &                &                &                &                                     &                  &                 \\ \hline
\multicolumn{1}{c|}{}                     & \multicolumn{1}{c|}{Types}                                                & OCG            & OCU            & OCI            & OCE            & RDG            & RDU            & RDI            & RDE            & GDG            & GDU            & GDI            & \multicolumn{1}{c|}{GDE}            & \multicolumn{2}{c}{MEAN}           \\ \hline
\multicolumn{1}{c|}{\multirow{9}{*}{FR}}  & \multicolumn{1}{c|}{Po2Po$_{MSE}$\citep{Rufaelm16}}       & \textbf{1.000} & \textbf{1.000} & \textbf{1.000} & 0.999          & \textbf{1.000} & 0.970          & \textbf{1.000} & \textbf{1.000} & \textbf{1.000} & \textbf{1.000} & 0.936          & \multicolumn{1}{c|}{0.975}          & \multicolumn{2}{c}{\textbf{0.995}} \\
\multicolumn{1}{c|}{}                     & \multicolumn{1}{c|}{Po2Po$_{HD}$\citep{Rufaelm16}}       & 1.000          & 1.000          & 1.000          & \textbf{1.000} & 0.999          & 0.964          & 0.999          & 0.994          & 1.000          & 0.999          & 0.966          & \multicolumn{1}{c|}{0.954}          & \multicolumn{2}{c}{0.995}          \\
\multicolumn{1}{c|}{}                     & \multicolumn{1}{c|}{Po2Po$_{PSNR}$\citep{Rufaelm16}}      & 0.998          & 1.000          & 1.000          & 1.000          & 1.000          & 0.939          & 1.000          & 1.000          & 1.000          & 1.000          & 0.915          & \multicolumn{1}{c|}{0.944}          & \multicolumn{2}{c}{0.992}          \\
\multicolumn{1}{c|}{}                     & \multicolumn{1}{c|}{Po2Pl$_{MSE}$\citep{Tiand17}}         & 0.997          & 1.000          & 1.000          & 0.999          & 1.000          & 0.949          & 0.977          & 1.000          & 1.000          & 0.960          & 0.943          & \multicolumn{1}{c|}{0.951}          & \multicolumn{2}{c}{0.991}          \\
\multicolumn{1}{c|}{}                     & \multicolumn{1}{c|}{Po2Pl$_{HD}$\citep{Tiand17}}         & 1.000          & 1.000          & 1.000          & 0.999          & 0.999          & 0.960          & 0.994          & 0.995          & 1.000          & 0.998          & 0.955          & \multicolumn{1}{c|}{0.942}          & \multicolumn{2}{c}{0.994}          \\
\multicolumn{1}{c|}{}                     & \multicolumn{1}{c|}{Po2Pl$_{PSNR} $\citep{Tiand17}}       & 0.999          & 1.000          & 1.000          & 1.000          & 1.000          & 0.905          & 0.995          & 1.000          & 1.000          & 0.987          & 0.899          & \multicolumn{1}{c|}{0.939}          & \multicolumn{2}{c}{0.989}          \\
\multicolumn{1}{c|}{}                     & \multicolumn{1}{c|}{Pl2Pl$_{MSE} $\citep{Alexioue18icme}} & 0.996          & 0.998          & 0.996          & 0.989          & 0.999          & 1.000          & 1.000          & 1.000          & 1.000          & 1.000          & \textbf{0.998} & \multicolumn{1}{c|}{\textbf{0.999}} & \multicolumn{2}{c}{0.992}          \\
\multicolumn{1}{c|}{}                     & \multicolumn{1}{c|}{Pl2Pl$_{HD}$\citep{Alexioue18icme}}  & 0.869          & 0.876          & 0.869          & 0.880          & 0.996          & 0.992          & 0.995          & 0.997          & 0.979          & 0.994          & 0.991          & \multicolumn{1}{c|}{0.986}          & \multicolumn{2}{c}{0.954}          \\
\multicolumn{1}{c|}{}                     & \multicolumn{1}{c|}{Pl2Pl$_{PSNR}$\citep{Alexioue18icme}} & 0.964          & 0.974          & 0.992          & 0.995          & 0.999          & \textbf{0.999} & 0.999          & 0.999          & 0.986          & 0.988          & 0.968          & \multicolumn{1}{c|}{0.992}          & \multicolumn{2}{c}{0.976}          \\
\multicolumn{1}{c|}{}                     & \multicolumn{1}{c|}{PCQM \citep{Meynetg20}}                                                 & 0.902          & 0.954          & 0.955          & 0.904          & 0.958          & 0.956          & 0.942          & 0.884          & 0.956          & 0.915          & 0.946          & \multicolumn{1}{c|}{0.950}          & \multicolumn{2}{c}{0.933}          \\
\multicolumn{1}{c|}{}                     & \multicolumn{1}{c|}{PointSSIM \citep{Alexioue20}}                                            & 0.915          & 1.000          & 0.998          & 1.000          & 0.997          & 1.000          & 1.000          & 1.000          & 0.902          & 1.000          & 1.000          & \multicolumn{1}{c|}{1.000}          & \multicolumn{2}{c}{0.990}          \\ \hline
\multicolumn{1}{c|}{\multirow{2}{*}{NR}}             & \multicolumn{1}{c|}{3DTA \citep{zhu20243dta}}                                                  & 0.903 &0.952 &0.899 &0.951 &0.898 &0.995 &0.928 &0.930 &0.924 &0.999 &0.917 & \multicolumn{1}{c|}{0.999} & \multicolumn{2}{c}{0.934}           \\
\multicolumn{1}{c|}{}                    & \multicolumn{1}{c|}{ours}                                               & 0.994  & 0.994  & 0.997  & 0.993  & 0.992  & 0.996  & 0.991  & 0.994  & 0.995  & 0.994  & 0.993   & \multicolumn{1}{c|}{0.994} & \multicolumn{2}{c}{0.994}           \\ \hline
\end{tabular}}
\end{table*}

The proposed LRL-GQA method aims to address the problem of no-reference geometry-only quality assessment for colorless point clouds.
However, as discussed in Section \ref{subsec:spcqa}, existing no-reference PCQA metrics are all designed to assess point clouds with both geometry and color attributes.
Therefore, no direct comparisons can be made with existing no-reference PCQA metrics.
For comparison fairness, three full-reference objective GQA metrics including Po2Po \citep{Rufaelm16}, Po2Pl \citep{Tiand17}, and Pl2Pl \citep{Alexioue18icme} are adopted in the experiments, with their detailed explanation provided in Section \ref{subsec:fr_method_details}.
And, three pooling strategies including mean squared error (MSE), HD, and geometric PSNR are introduced to obtain the overall quality distortion \citep{Javaheri21, zhoul22}.
In addition, two full-reference metrics, point cloud structural similarity metric 
(PointSSIM) \citep{Alexioue20} and point cloud quality
metric (PCQM) \citep{Meynetg20}, along with one no-reference metric, 3D point cloud quality assessment with twin attention (3DTA) \citep{zhu20243dta}, are employed for the evaluation of colored point clouds.
PointSSIM assesses SSIM by comparing feature mappings associated with the reference and distorted point clouds in local neighborhoods, while PCQM provides an evaluation considering the geometric structure, color fidelity, and illumination consistency of point clouds. 
On the other hand, 3DTA leverages twin attention to enhance the perceptual capabilities of feature representations.
Specifically, we have fine-tuned the parameters of PCQM to perform better on the LRL-PMOS dataset. 
Regarding the issue of no-reference metrics targeting colored point clouds, we uniformly define the point clouds of the LRL dataset as gray for comparison.
All the full-reference and no-reference results are obtained using the source code released by the authors.
Thus, a total of 12 metrics are tested on the proposed datasets: Po2Po$_{MSE}$ \citep{Rufaelm16}, Po2Po$_{HD}$ \citep{Rufaelm16}, Po2Po$_{PSNR}$ \citep{Rufaelm16}, Po2Pl$_{MSE}$ \citep{Tiand17}, Po2Pl$_{HD}$ \citep{Tiand17}, Po2Pl$_{PSNR}$ \citep{Tiand17}, Pl2Pl$_{MSE}$ \citep{Alexioue18icme}, Pl2Pl$_{HD}$ \citep{Alexioue18icme}, Pl2Pl$_{PSNR}$ \citep{Alexioue18icme}, PCQM \citep{Meynetg20}, PointSSIM \citep{Alexioue20} and 3DTA \citep{zhu20243dta}.

\subsubsection{Evaluation Criteria}

To evaluate the ranking performance of the proposed LRL-GQA method, the normalized discounted cumulative gain (NDCG) \citep{Qint10, ZehlikeM22} is used.
The NDCG, which is widely adopted in the information retrieval field, compares the ranking generated by the proposed method to the ground-truth ranking (also known as the ideal ranking).
The NDCG is calculated as the ratio of the discounted cumulative gain (DCG) to the ideal DCG (IDCG), as shown in Eq.~\eqref{eq:ndcg} and Eq.~\eqref{eq:dcg}:
\begin{equation}
	\label{eq:ndcg}
	\mathrm{NDCG} = \frac{\mathrm{DCG}}{\mathrm{IDCG}} =  \frac{\mathrm{DCG}}{max (\mathrm{DCG})},
\end{equation}
where 
\begin{equation}
	\label{eq:dcg}
	\mathrm{DCG} = \sum_{i=1}^{k}\frac{rel(i)}{log_2(i + 1)},
\end{equation}
$k$ is the size of the ranking list ($k = 11$ in this paper),  $rel(i) = 0.5 + 0.5 \times (k - i) / (k - 1)$ is the graded relevance of the result at position $i$ in this paper .
It can be seen that the DCG weighs each relevance score based on its position, which indicates that the recommendations at the top get a higher weight while the relevance of those at the bottom get a lower weight.
And, the IDCG denotes the maximum of DCG which can be obtained when the samples are ranked in the ideal order.

To assess the scoring performance of the proposed LRL-GQA method, four common evaluation metrics including root mean square error (RMSE), Pearson linear correlation coefficient (PLCC) \citep{Sedgwicke12}, Kendall rank correlation coefficient (KRCC) \citep{liuq21tcsvt}, and Spearman rank correlation coefficient (SRCC) \citep{Yangq21} are employed.

\subsection{Quality Ranking}

The ranking performance of the proposed LRL-GQA method are tested on the LRL dataset.
Specifically, 90\% of the reference point clouds and their distorted versions from the LRL dataset are selected for training while the remaining 10\% are for testing.
Overall, there are a total of $200 \times 0.9 \times 26 = 4680$ training lists and $200 \times 0.1 \times 26 = 520$ testing lists.
For the selected three full-reference quality evaluation methods, the quality indexes of point clouds in each testing list are first calculated.
Then, the point clouds in each testing list are ranked according to their predicted indexes.
Table \ref{tab:NDCG10} compares the ranking accuracy of different methods, including both full-reference (FR) and no-reference (NR) methods,  for each type of distortion on the entire testing lists in terms of NDCG.
It can be seen that the lowest NDCG of the proposed LRL-GQA method is greater than 0.992.
And, its overall performance is 0.994, which is comparable with selected full-reference metrics and outperforms the compared no-reference metrics.
Additionally, Fig. \ref{fig:existing_techniques_visually} provides a visual comparison and computing consumption between the proposed approach and FR methods.
Specifically, we selected five evenly spaced levels from a ten-level distortion to illustrate the scores calculated by different algorithms.
It can be observed that the proposed no-reference LRL-GQA method is able to assign different scores to point clouds of varying quality.

\begin{table}[!t]
	\caption{NDCG scores  of the proposed LRL-GQA method on G-PCD dataset. Higher values imply better outcomes.}
	\label{tab:LRLNDCG_GPCD}
	\centering

 	\begin{tabular}{c|l|ll}
		\hline
		Types                            & Methods     & GN    & OC    \\ \hline
		\multirow{11}{*}{FR} 
        & Po2Po$_{MSE}$\citep{Rufaelm16} & 0.994 & 0.988 \\
		& Po2Po$_{HD}$\citep{Rufaelm16} & 0.998 & 0.996 \\
		& Po2Po$_{PSNR}$\citep{Rufaelm16} & 0.996 & 0.988 \\
		& Po2Pl$_{MSE}$\citep{Tiand17} & 0.999 & 0.996 \\
		& Po2Pl$_{HD}$\citep{Tiand17} & 1.000 & 1.000 \\
		& Po2Pl$_{PSNR}$\citep{Tiand17} & 0.999 & 1.000 \\
		& Pl2Pl$_{MSE}$\citep{Alexioue18icme} & 0.999 & 0.997 \\
		& Pl2Pl$_{HD}$\citep{Alexioue18icme} & 0.998 & 0.999 \\
		& Pl2Pl$_{PSNR}$\citep{Alexioue18icme} & 0.975 & 0.889 \\
		& PCQM\citep{Meynetg20}        & 1.000     & 1.000     \\
		& PointSSIM\citep{Alexioue20}   & 1.000     & 1.000     \\
    \hline
		\multirow{2}{*}{NR}    
        & 3DTA\citep{zhu20243dta}        & 0.928     & 0.947\\
		& Ours        & 0.982 & 0.982    \\
  \hline
\end{tabular}
\end{table}

To further evaluate the generalization of the proposed LRL-GQA method, each reference point cloud of the testing lists is degraded by same distortion types at 20 distortion levels.
That is, each testing list consists of 1 reference point cloud and 20 distorted versions with the same distortion type.
The proposed LRL-GQA method, which is trained on the training lists with only 10 distortion levels, is then tested on the new testing lists.
Table \ref{tab:NDCG20} lists the NDCG of different methods for each type of distortion with 20 levels.
It can be observed that the proposed LRL-GQA method can still obtain high ranking accuracy on the testing lists with 20 distortion levels.
Therefore, it is qualified for assessing or ranking degraded point clouds with more subtle differences.
We also conduct an experiment on the G-PCD dataset, a small pure geometry dataset with GN and OC distortions, to evaluate the model's generalization performance.
It can be seen from Table \ref{tab:LRLNDCG_GPCD} that the proposed LRL-GQA method achieves high ranking accuracy even on previously unseen data. 

\begin{table}[!t]
	\centering
	\caption{Absolute quality score prediction performance of the fine-tuned GQANet on the LRL-PMOS dataset with pseudo MOS. The red, green, and blue colors indicate the first three best results, respectively.}
	\label{tab:PMOSresult}
	\resizebox{0.48\textwidth}{!}{
	\begin{tabular}{@{}c|l|llll@{}}
		\hline
		\multicolumn{1}{c|}{Types}  & Methods   & RMSE & PLCC & KRCC & SRCC \\ \hline
		\multirow{9}{*}{FR} 
            & Po2Po$_{MSE} $\citep{Rufaelm16} & 0.149 & 0.803 & 0.573 & 0.742 \\
		& Po2Po$_{HD}$\citep{Rufaelm16} & 0.155 & 0.789 & 0.563 & 0.739 \\
		& Po2Po$_{PSNR}$\citep{Rufaelm16}      & 0.153 & 0.794 & 0.570& 0.743 \\
		& Po2Pl$_{MSE} $\citep{Tiand17}      & 0.128 & 0.860 & 0.641 & 0.805 \\
		& Po2Pl$_{HD}$\citep{Tiand17} & 0.149 & 0.807& 0.581 & 0.759 \\
		& Po2Pl$_{PSNR}$\citep{Tiand17}     & 0.130 & 0.856& 0.637 & 0.507 \\
		& Pl2Pl$_{MSE} $\citep{Alexioue18icme}      & \textcolor{red}{0.016} & \textcolor{red}{0.998} &\textcolor{red}{0.972} & \textcolor{red}{0.998} \\
		& Pl2Pl$_{HD}$\citep{Alexioue18icme} & 0.158 & 0.779 & 0.619 & 0.794 \\
		& Pl2Pl$_{PSNR} $\citep{Alexioue18icme}     & \textcolor{green}{0.082} & \textcolor{green}{0.945}  & \textcolor{green}{0.805}  & \textcolor{green}{0.949} \\
 & PCQM\citep{Meynetg20}     & 0.152 & 0.359 & 0.277 & 0.367 \\
 & PointSSIM\citep{Alexioue20}    & 0.166 & 0.681& 0.449 & 0.593 \\
  \hline
		\multirow{2}{*}{NR} 
           & 3DTA \citep{zhu20243dta} & 0.173 & 0.731 & 0.483 & 0.668  \\ 
            & Ours &\textcolor{blue}{0.077} & \textcolor{blue}{0.927}  & \textcolor{blue}{0.754} & \textcolor{blue}{0.841} \\ \hline
	\end{tabular}
	}
\end{table}

\begin{figure}[!t]
	\centering
	\includegraphics[width=0.5\textwidth]{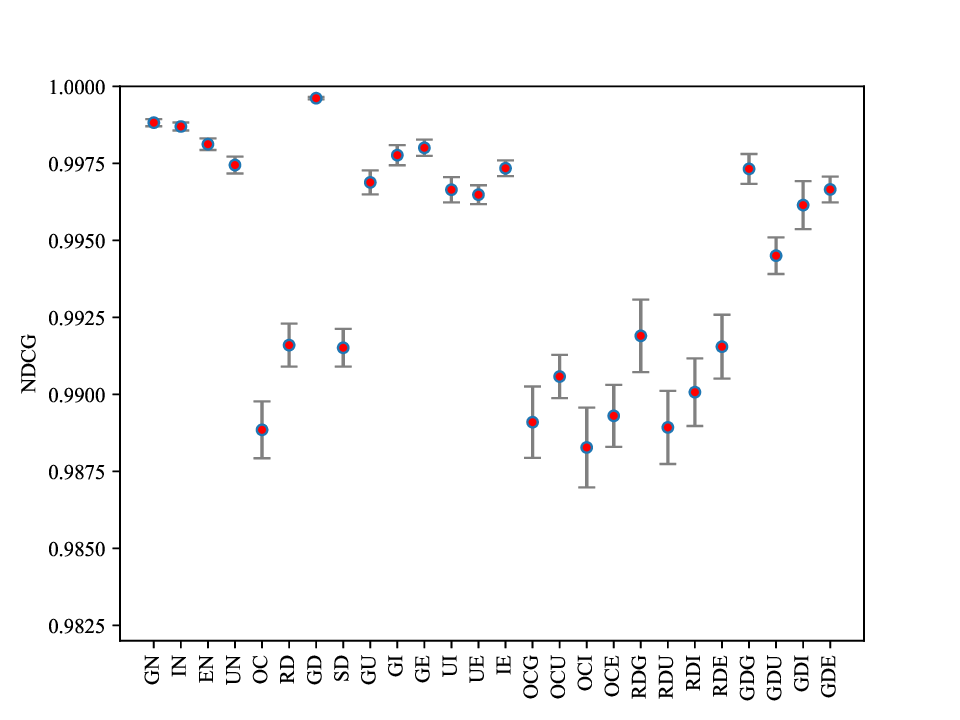}
	\caption{ Performance and stability analysis under various degradation types. }
	\label{fig:error_bar}
\end{figure}

Furthermore, we investigate the performance of the proposed LRL-GQA across distinct noise conditions in terms of both mean and variance, where the mean reflects accuracy and the variance represents stability.
From Fig. \ref{fig:error_bar}, it can be observed that both accuracy and stability decrease under compression distortions (OC, RD) and combination noise types which include compression distortions.
This is because that compression operations (OC, RD) reduce the point count, resulting in the loss of fine details, which subsequently weaken the network's ability to capture features. 

\begin{table}[!t]
	\caption{Computation and storage analysis on the G-PCD dataset, including both traditional methods (TR) and deep learning-based methods (DL).}
	\label{tab:computation_storage}
	\centering

	\begin{tabular}{l|l|rr}
		\hline
		Types                          & Method    & Time    & Memory     \\ \hline
		\multirow{5}{*}{TR}   
& Po2Po\citep{Rufaelm16}     & 62.4 ms   & 11.51MB    \\
& Po2Pl\citep{Tiand17}     & 72.2 ms   & 13.48 MB    \\
& Pl2Pl\citep{Alexioue18icme}     & 80.6 ms   & 14.20 MB    \\
& PCQM\citep{Meynetg20}      & 57.1 ms   & 83.69 MB    \\
& PointSSIM\citep{Alexioue20} & 20122 ms & 240.64 MB  \\ \hline
		\multirow{2}{*}{DL} & 3DTA \citep{zhu20243dta}     
& 2654.8 ms   & 7619.25 MB \\
& Ours      & 343.6 ms  & 935.82 MB  \\ \hline
	\end{tabular}
\end{table}

\begin{table*}[!t]
	\centering
 \setlength{\tabcolsep}{4pt}
	\caption{NDCG scores  of the proposed LRL-GQA method with and without patch generation (PG) in the GQANet on the LRL Dataset. Higher values imply better outcomes.}
        \label{tab:patchLRLNDCG}
        \resizebox{\textwidth}{!}{
       \begin{tabular}{cccccccccccccccc}
       \hline                   & \multicolumn{1}{c|}{Types}      & GN    & IN    & EN    & UN    & OC    & RD    & GD    & SD    & GU    & GI    & GE              &UI           & UE           & IE         \\ \hline
 & \multicolumn{1}{c|}{LRL-GQA (without PG)} & \textbf{0.999} & \textbf{0.998} & \textbf{0.998} & \textbf{0.999} & \textbf{0.998} & \textbf{0.998}  & \textbf{0.999}  &\textbf{0.998}& \textbf{0.998}&\textbf{0.998}&\textbf{0.997}&\textbf{0.998}&\textbf{0.999}&\textbf{0.998}  \\
 & \multicolumn{1}{c|}{LRL-GQA (with PG)}   &0.993& 0.994& 0.997 & 0.996 &0.996 & 0.994 & 0.995& 0.993 & 0.992 &0.997&0.994&0.995&0.994&0.993    \\\hline
     &                                &       &       &       &       &       &       &       &       &       &       &                            &              &            \\\hline
 & \multicolumn{1}{c|}{Types}     & OCG    & OCU    & OCI    & OCE    & RDG    & RDU    & RDI    & RDE    & GDG    & GDU   &GDI & \multicolumn{1}{c|}{GDE}    & \multicolumn{2}{c}{MEAN}  \\ \hline
 & \multicolumn{1}{c|}{LRL-GQA (without PG)} & 0.992 & 0.993 & \textbf{0.993} & 0.992 & \textbf{0.997} & 0.994 & \textbf{0.995} & 0.994 & \textbf{0.997} & \textbf{0.996} & \textbf{0.997}  & \multicolumn{1}{c|}{\textbf{0.995}} & \multicolumn{2}{c}{\textbf{0.996}} \\
 & \multicolumn{1}{c|}{LRL-GQA (with PG)}   & \textbf{0.993} & \textbf{0.993} & 0.992 & \textbf{0.994} & 0.995 & \textbf{0.994} &0.993 & \textbf{0.994} & 0.994   & 0.993 & 0.995 & \multicolumn{1}{c|}{0.992} & \multicolumn{2}{c}{0.994} \\

\hline
\end{tabular}}
\end{table*}

\begin{table}[!t]
	\centering
	\caption{Performance of the proposed GQANet with and without patch generation (PG) on the LRL-PMOS Dataset.}
	\begin{tabular}{@{}c|cccc@{}}
		\hline
		\multicolumn{1}{c|}{Methods} & RMSE & PLCC & KRCC & SRCC \\ \hline
		GQANet (without PG) & 0.054& 0.981& 0.874& 0.979\\ 
		GQANet (with PG) & \textbf{0.031}& \textbf{0.995}& \textbf{0.932}& \textbf{0.994}\\ \hline
	\end{tabular}
	\label{tab:patchLRLScore}
\end{table}

\begin{table}[!ht]
	\centering
	\caption{Performance of the proposed GQANet with and without patch generation (PG) on the G-PCD Dataset.}
	\begin{tabular}{@{}c|cccc@{}}
		\hline
		\multicolumn{1}{c|}{Methods} & RMSE & PLCC & KRCC & SRCC \\ \hline
		GQANet (without PG) & 0.330& 0.968& 0.894& 0.968\\ 
		GQANet (with PG) & \textbf{0.141}& \textbf{0.994}& \textbf{0.941}& \textbf{0.991}\\ \hline
	\end{tabular}
	\label{tab:patchGPCDScore}
\end{table}

\begin{table*}[!t]
	\centering
  \setlength{\tabcolsep}{4pt}
	\caption{NDCG scores  of the proposed LRL-GQA method with different feature extraction networks on the LRL dataset. Higher values imply better outcomes.}
        \label{tab:pointnet}
        \resizebox{\textwidth}{!}{
       \begin{tabular}{cccccccccccccccc}
       \hline
              & \multicolumn{1}{c|}{Types}      & GN    & IN    & EN    & UN    & OC    & RD    & GD    & SD    & GU    & GI    & GE              &UI           & UE           & IE         \\ \hline
 & \multicolumn{1}{c|}{LRL-GQA (with PointNet)} & 0.990 & 0.992 & 0.992 & 0.992 & 0.989 & 0.990 & 0.991 &0.985& 0.988&0.992&0.989&0.992&0.990&0.985 \\
               & \multicolumn{1}{c|}{LRL-GQA (with DGCNN)}   & 0.993 & \textbf{0.995} & 0.996 & 0.996 & 0.994 & \textbf{0.995} & 0.995 & 0.990 &0.989 &0.996 &\textbf{0.996}& 0.995 &0.994 & 0.993 \\
               & \multicolumn{1}{c|}{LRL-GQA (with PCT)}   & \textbf{0.995} & 0.992 & 0.994 & 0.991 & 0.994 & 0.987 & 0.986 & 0.981&0.986&0.988&0.983&0.983&0.988&0.978\\
                & \multicolumn{1}{c|}{LRL-GQA (with MPFE)}   & 0.993 & 0.994& \textbf{0.997} & \textbf{0.996} & \textbf{0.996} & 0.994 & \textbf{0.995} & \textbf{0.993} & \textbf{0.992} & \textbf{0.997} & 0.994 & \textbf{0.995} & \textbf{0.994} & \textbf{0.993} \\
\hline
                                         &                                &       &       &       &       &       &       &       &       &       &       &                            &              &            \\\hline
                & \multicolumn{1}{c|}{Types}     & OCG    & OCU    & OCI    & OCE    & RDG    & RDU    & RDI    & RDE    & GDG    & GDU   &GDI & \multicolumn{1}{c|}{GDE}    & \multicolumn{2}{c}{MEAN}  \\ \hline
 & \multicolumn{1}{c|}{LRL-GQA (with PointNet)} &0.987 & 0.987 & 0.989 & 0.989 & 0.992 & 0.988 & 0.987 & 0.992 & 0.991 & 0.988 & 0.987   & \multicolumn{1}{c|}{0.989} & \multicolumn{2}{c}{0.989} \\
              & \multicolumn{1}{c|}{LRL-GQA (with DGCNN)}   & 0.992 & 0.992 & \textbf{0.995} & 0.993 & 0.995 & 0.993 & 0.991 & \textbf{0.996} & \textbf{0.996} & 0.993 &0.995   & \multicolumn{1}{c|}{0.992} & \multicolumn{2}{c}{0.994} \\
              & \multicolumn{1}{c|}{LRL-GQA (with PCT)}   & 0.978 & 0.981 & 0.988 & 0.994 & 0.991 & 0.990 & 0.975 & 0.990 & 0.989 & 0.973 &\textbf{0.997}   & \multicolumn{1}{c|}{0.983} & \multicolumn{2}{c}{0.987} \\
                & \multicolumn{1}{c|}{LRL-GQA (with MPFE)}  &\textbf{0.993}& \textbf{0.993} & 0.992 & \textbf{0.994} & \textbf{0.995} & \textbf{0.994} & \textbf{0.993} & 0.994 & 0.994   & \textbf{0.993} & 0.995 & \multicolumn{1}{c|}{\textbf{0.992}} & \multicolumn{2}{c}{\textbf{0.994}}
 \\

\hline
\end{tabular}}
\end{table*}

\begin{table*}[!t]
	\centering
 \setlength{\tabcolsep}{4pt}
	\caption{NDCG scores  of the proposed LRL-GQA method with and without the weighted quality index calculation module on the LRL dataset. Higher values imply better outcomes.}
        \label{tab:weight}
        \resizebox{\textwidth}{!}{
       \begin{tabular}{cccccccccccccccc}
       \hline
                   & \multicolumn{1}{c|}{Types}      & GN    & IN    & EN    & UN    & OC    & RD    & GD    & SD    & GU    & GI    & GE              &UI           & UE           & IE         \\ \hline
 & \multicolumn{1}{c|}{LRL-GQA (without weights)} & 0.933& 0.931 & 0.935 & 0.932 & 0.930 & 0.928 & 0.930 &0.925& 0.931 &0.929&0.927&0.931&0.930&0.922 \\
                   & \multicolumn{1}{c|}{LRL-GQA (with weights)}  &\textbf{0.993}& \textbf{0.994}& \textbf{0.997} & \textbf{0.996} &\textbf{0.996} & \textbf{0.994} & \textbf{0.995}& \textbf{0.993} & \textbf{0.992} &\textbf{0.997}&\textbf{0.994}&\textbf{0.995}&\textbf{0.994}&\textbf{0.993} \\\hline
                                         &                                &       &       &       &       &       &       &       &       &       &       &                            &              &            \\\hline
        & \multicolumn{1}{c|}{Types}     & OCG    & OCU    & OCI    & OCE    & RDG    & RDU    & RDI    & RDE    & GDG    & GDU   &GDI & \multicolumn{1}{c|}{GDE}    & \multicolumn{2}{c}{MEAN}  \\ \hline
 & \multicolumn{1}{c|}{LRL-GQA (without weights)} & 0.919&0.920&0.912&0.920&0.925&0.927&0.930&0.927&0.929&0.927&0.921  & \multicolumn{1}{c|}{0.922} & \multicolumn{2}{c}{0.925} \\
    & \multicolumn{1}{c|}{LRL-GQA (with weights)}   & \textbf{0.993} & \textbf{0.993} & \textbf{0.992} & \textbf{0.994} & \textbf{0.995} & \textbf{0.994} & \textbf{0.993} & \textbf{0.994} & \textbf{0.994} & \textbf{0.993} & \textbf{0.995}   & \multicolumn{1}{c|}{\textbf{0.992}} & \multicolumn{2}{c}{\textbf{0.994}} \\

\hline
\end{tabular}}
\end{table*}

\subsection{Quality Score Calculation}
\label{subsec:score}

A new subjective dataset with pseudo MOS, called LRL-PMOS, is designed to validate the scoring performance of the proposed GQANet, since there are currently no large-scale colorless PCQA datasets with subjective MOS where all point clouds are degraded by only geometric distortions.
To this end, the concept of pseudo MOS which has been validated and used in both image and point cloud quality assessments \citep{Wujj20,oufz21,Liuyp22} is introduced to generate pseudo MOS.
To be specific, for each reference point cloud in the LRL Dataset, the degree of distortion is measured by the angular similarity \citep{Alexioue18icme}, which is a kind of high-performance Pl2Pl distance.
This metric estimates the similarity between the reference and degraded point clouds by modeling the surfaces in both point clouds as planes.
In addition, its output is used directly as a subjective score between [0.0, 1.0].
Therefore, the LRL-PMOS dataset contains $200 \times 26 \times 10 = 52200$ geometrically distorted samples with pseudo MOS.

\begin{table*}[!t]
	\centering
 \setlength{\tabcolsep}{6pt}
	\caption{NDCG scores of the proposed LRL-GQA method of different shapes on the LRL Dataset. Higher values imply better outcomes.}
        \label{tab:shapeLRLNDCG}
        \resizebox{\textwidth}{!}{
       \begin{tabular}{cccccccccccccccc}
       \hline                   & \multicolumn{1}{c|}{Types}      & GN    & IN    & EN    & UN    & OC    & RD    & GD    & SD    & GU    & GI    & GE              &UI           & UE           & IE         \\ \hline
 & \multicolumn{1}{c|}{Regular Objects} & 0.998& 0.997& 0.999& 1.000& 0.998& 0.959& 0.933& 0.985& 1.000& 1.000& 0.999& 1.000& 1.000& 0.999   \\
 & \multicolumn{1}{c|}{Irregular Objects}   & 0.996& 0.995& 0.998& 0.998& 0.996& 0.957& 0.919& 0.980& 1.000& 0.999& 0.998& 0.999& 0.999& 0.998    \\\hline
     &                                &       &       &       &       &       &       &       &       &       &       &                            &              &            \\\hline
 & \multicolumn{1}{c|}{Types}     & OCG    & OCU    & OCI    & OCE    & RDG    & RDU    & RDI    & RDE    & GDG    & GDU   &GDI & \multicolumn{1}{c|}{GDE}    & \multicolumn{2}{c}{MEAN}  \\ \hline
 & \multicolumn{1}{c|}{Regular Objects}& 1.000 & 1.000& 1.000& 1.000& 0.997& 0.999& 0.998& 0.998& 0.998& 1.000& 0.999& \multicolumn{1}{c|}{0.999} & \multicolumn{2}{c}{0.994}  \\
 & \multicolumn{1}{c|}{Irregular Objects}& 0.999 & 1.000& 0.999& 0.999& 0.989& 0.996& 0.994& 0.994& 0.990& 0.995& 0.994&  \multicolumn{1}{c|}{0.991} & \multicolumn{2}{c}{0.991} \\

\hline
\end{tabular}}
\end{table*}

Theoretically, the reliability of generated pseudo MOS can be verified via calculating the correlation between the pseudo and subjective MOS \citep{Alexioue18icme,Wujj20,Liuyp22}.
However, this paper does not justify the accuracy of the computed pseudo MOS in this way for the following reasons: 
(1) It has been demonstrated in \citep{Alexioue18icme} that the reliability and accuracy of the adopted angular similarity is consistent with subjective quality assessment scores under certain types of geometry distortions, including Gaussian noise and compression-like artifacts.
(2) It is quite difficult to collect subjective MOS for a large number of point clouds through time-consuming and expensive subjective experiments under strict control conditions.
(3) The NDCG performance shown in Table \ref{tab:NDCG10} and Table \ref{tab:NDCG20} also demonstrates that the generated pseudo MOS can objectively reflect the relative quality relationship among different degraded point clouds associated with the same distortion.

To evaluate the scoring performance, all point clouds from the synthetic LRL-PMOS dataset are randomly divided into two non-overlapping subsets: 90\% for training and 10\% for testing.
The pre-trained GQANet is firstly fine-tuned on the training set and then tested on the testing set. 
Its output is utilized directly as absolute quality scores.
To compare the performance of the GQANet with existing full-reference geometry-only metrics, the RMSE, PLCC, SRCC, and KRCC are introduced. 
Table \ref{tab:PMOSresult} gives the comparison results on the LRL-PMOS dataset.
It can be seen that, given a dataset with ground-truth MOS, the fine-tuned GQANet has a good ability to predict absolute quality scores. 
The GQANet achieves competitive prediction performance and outperforms the majority of existing full-reference evaluation metrics as a no-reference geometry-only quality assessment method. 
Additionally, it can be found that the selected three metrics for colored point clouds (PointSSIM \citep{Alexioue20}, PCQM \citep{Meynetg20}, and 3DTA \citep{zhu20243dta}) perform worse than  geometry-only metrics.
These results also indicate that the color information plays an important role in quality evaluation and existing PCQA metrics for colored point clouds cannot be directly applied to the GQA task.
Therefore, it is anticipated that the GQANet will perform well on large GQA datasets with subjective MOS, if any, due to its advantageous multi-scale quality-aware feature extraction capability.

\subsection{Computation and Storage Analysis}
\label{subsec:computation_storage}

To evaluate the computational efficiency, we measure the average inference time and memory consumption per point cloud on the entire G-PCD dataset, which consists of a total of 50 samples.
The algorithms to be evaluated include both traditional methods (Po2Po\citep{Rufaelm16}, Po2Pl\citep{Tiand17}, and Pl2Pl\citep{Alexioue18icme}) and deep learning-based methods (3DTA \citep{zhu20243dta} and the proposed GQANet).
Among the Po2Po\citep{Rufaelm16}, Po2Pl\citep{Tiand17}, and Pl2Pl\citep{Alexioue18icme} methods, the metrics MSE, HD, and PSNR differ only in the way the final distance is calculated.
Therefore, these metrics have similar computational overheads.

From Table \ref{tab:computation_storage}, it can be observed that deep learning-based methods require significantly more memory and longer computation time compared to traditional methods on the Intel i9-13900K CPU.
Note that deep learning-based methods involve extensive matrix operations, which are not the strength of CPU.
Therefore, we also use a RTX 4090 GPU to accelerate both 3DTA and the proposed GQANet.
The 3DTA requires 91.4 ms to process a single point cloud in G-PCD dataset, while the proposed method needs just 45.6 ms.
 
\subsection{Ablation Experiments}
Three comparative experiments are designed to verify the contribution of each component in the proposed LRL-GQA network in this section.

\subsubsection{Effectiveness of Patch Generation}

To testify the effectiveness of the patch generation, the GQANet is simplified and contains only the MPFE and patch quality index calculation module.
The patch quality weight calculation and weighted model quality index calculation modules are removed, as the MPFE takes the entire point cloud as input.
The output of the patch quality index calculation module is directly used as the model quality index.

Table \ref{tab:patchLRLNDCG} gives the ranking performance on the LRL Dataset.
It can be observed that the LRL-GQA with patch generation achieves comparable ranking results with the one without patch generation.
In addition, Table \ref{tab:patchLRLScore} and \ref{tab:patchGPCDScore} present the scoring performance on the LRL-PMOS dataset and G-PCD dataset \citep{Alexioue17}, respectively.
The G-PCD is the largest available colorless dataset with subjective ground-truth MOS, containing 5 colorless reference point clouds and 40 samples distorted by octree-puring and Gaussian noise.
Note that, all the reference point clouds and their distorted samples in the G-PCD dataset are both used for fine-tuning and testing, because of the limited number of samples.
It can be seen from Table \ref{tab:patchLRLScore} and \ref{tab:patchGPCDScore} that the overall scoring performance of GQANet with patch generation is the best on LRL-PMOS dataset and G-PCD dataset, but is not yet perfect.
This is because that the distortions included in existing datasets are synthetic and uniformly distributed over the entire point clouds.
Furthermore, the entire LRL-GQA network is not trained in an end-to-end mode, as discussed in Section \ref{subsec:Training-tuning}.
Consequently, the advantages of the patch generation cannot be fully utilized.

\subsubsection{Effectiveness of Multi-Scale Patch Feature Extraction}

To demonstrate the effectiveness of the MPFE module, the GQANet employs the PointNet \citep{qi2017pointnet}, labeled directed graphs convolutional neural network (DGCNN) \citep{Wangy19} and point cloud transformer (PCT) \citep{Guomh21} as the feature extraction backbone for testing, respectively. 
Table \ref{tab:pointnet} summarizes the ranking accuracy performance of the proposed method before and after the MPFE was removed. 
It can be seen that the overall performance of the proposed LRL-GQA with MPFE is the best.
This demonstrates the essential contributions made by the MPFE.

\subsubsection{Effectiveness of Patch Quality Weight Calculation} 

\begin{figure}[t]
\centering  
\begin{subfigure}{0.4\linewidth}
    \centering
    \includegraphics[width=\linewidth]{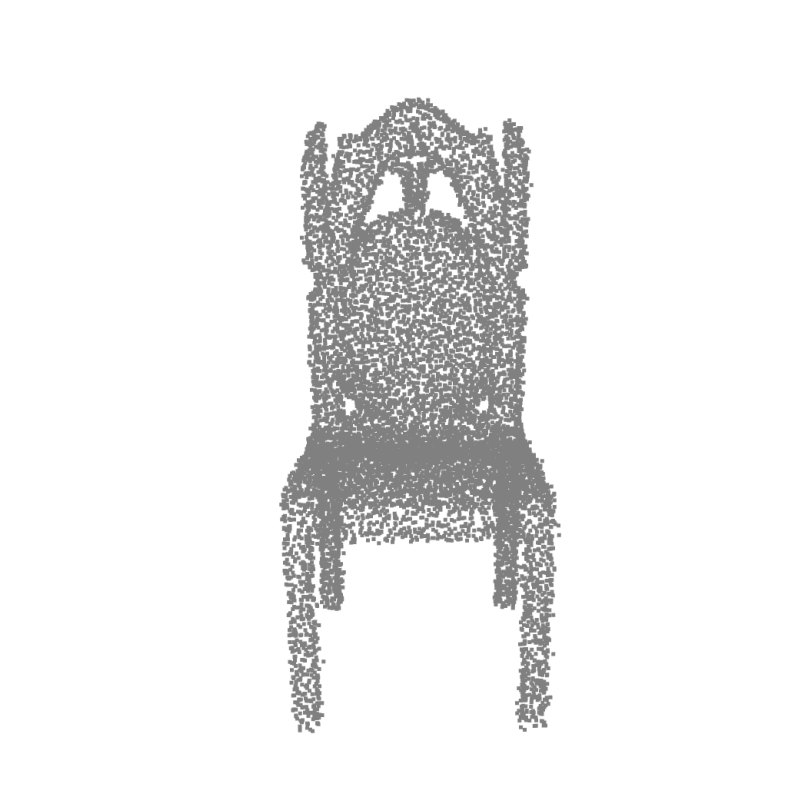}
    \caption{Original chair}
    \label{fig:original_chair}
\end{subfigure}
\begin{subfigure}{0.4\linewidth}
    \centering
    \includegraphics[width=\linewidth]{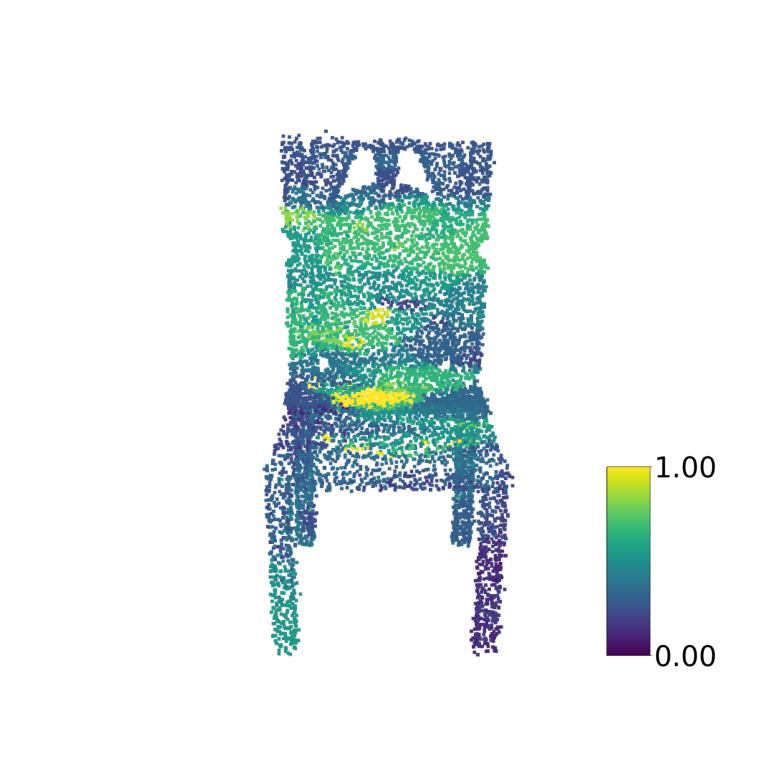}
    \caption{Colored chair}
    \label{fig:colored_chair}
\end{subfigure}
\caption{Visualization of weights of different patches of a chair.}
\label{tab:colormap}
\end{figure}

To evaluate the effectiveness of the patch quality weight calculation module in the GQANet, the proposed LRL-GQA method assigns equal weights to all patches and averages normalized patch scores.
The ranking accuracy of the LRL-GQA with and without the weighted quality index calculation module is listed in Table \ref{tab:weight}.
It can be observed that the patch quality weight calculation module significantly improves the ranking accuracy. 
This indicates the indispensable contributions of the patch quality weight calculation module to the proposed method.
From Fig.\ref{tab:colormap}, it can be seen that different patches are assigned different weights. 
In areas with flat surfaces, the weight is relatively low and the color tends to be cooler. 
Contrastingly, in regions abundant with details such as the junction between the chair back and seat, the weighting is increased, resulting in warmer colors.

\subsubsection{Effectiveness of Shape Regularity} 

In order to investigate the impact of regular and irregular point clouds on the performance, we conduct experiments on the LRL dataset by dividing the point clouds into regular and irregular groups. 
Experimental results are shown in Table \ref{tab:shapeLRLNDCG}.
It can be observed that the effect of point cloud shape regularity on the results is minimal.
However, regular point clouds demonstrate slightly higher performance compared to irregular ones. 
We attribute this to the greater difficulty in predicting the list ranking of irregular point clouds compared to regular ones.

\section{Conclusion}
\label{sec:Conclusion}

This paper tackles the problem of geometry-only quality assessment for distorted colorless point clouds based on list-wise rank learning, called LRL-GQA.
The no-reference GQA is formulated as a ranking problem and the point cloud lists associated with the same type distortion at various levels are used as learning instances.
The LRL-GQA first employs a specifically designed GQANet to encode intrinsic multi-scale patch-wise geometric features in order to predict a quality index for every input point cloud.
Then, a list-wise ranking approach is developed to learn a quality predictor by preserving the subjective ranking of degraded point clouds generated from the same source point cloud.
Due to the lack of publicly available GQA datasets, a large-scale list-wise dataset for the rank learning-based GQA task is built.
Experiments validate the ranking performance of the proposed LRL-GQA and the scoring performance of the GQANet.
In the future, we will further investigate the problem of geometry-only quality assessment for colored point clouds.

\section{Acknowledgments}

The authors sincerely acknowledge the anonymous reviewers for their insights and comments to further improve the quality of the manuscript.

\bibliographystyle{elsarticle-num} 
\bibliography{LA}

\end{document}